%% file: main.tex
\documentclass[runningheads]{llncs}

\usepackage[mobile]{eccv}

\usepackage{eccvabbrv}

\usepackage{subcaption}
\usepackage{graphicx}
\usepackage{booktabs}
\usepackage{tikz}
\usepackage{comment}
\usepackage{amsmath,amssymb} % define this before the line numbering.
\usepackage{color}
\usepackage{multirow}
\usepackage{colortbl}
\usepackage{multibib}

\usepackage[accsupp]{axessibility}  % Improves PDF readability for those with disabilities.

\usepackage{hyperref}

\usepackage{orcidlink}

\newcites{SM}{Supplementary References}

\begin{document}

% ---------------------------------------------------------------

\title{Open-Set Recognition in the Age of Vision-Language Models} 

\author{Dimity Miller \and
Niko S\"underhauf \and
Alex Kenna \and
Keita Mason
}

\authorrunning{D.~Miller et al.}

\institute{
    Queensland University of Technology, Brisbane, Australia%
    \footnote{The authors acknowledge ongoing support from the QUT Centre for Robotics. \\Evaluation code is publicly available at \href{http://github.com/dimitymiller/openset\_vlms}{\url{github.com/dimitymiller/openset\_vlms}}}\\
\email{\{d24.miller, niko.suenderhauf\}@qut.edu.au}}

\maketitle
\input{mathstuff}

\begin{abstract}
Are vision-language models (VLMs) for open-vocabulary perception inherently open-set models because they are trained on internet-scale datasets? We answer this question with a clear \emph{no} -- VLMs introduce closed-set assumptions via their finite query set, making them vulnerable to open-set conditions. We systematically evaluate VLMs for open-set recognition and find they frequently misclassify objects not contained in their query set, leading to alarmingly low precision when tuned for high recall and vice versa.
We show that naively increasing the size of the query set to contain more and more classes does \emph{not} mitigate this problem, but instead causes diminishing task performance \emph{and} open-set performance.
We establish a revised definition of the open-set problem for the age of VLMs, define a new benchmark and evaluation protocol to facilitate standardised evaluation and research in this important area, and evaluate promising baseline approaches based on predictive uncertainty and dedicated negative embeddings on a range of open-vocabulary VLM classifiers and object detectors.
\end{abstract}

\input{sections/introduction}

\input{sections/background}

\input{sections/problem}

\input{sections/evaluate}

\input{sections/experiments}

\input{sections/discussion}


\bibliographystyle{splncs04}
\bibliography{refs}

\input{sections/supplementary.tex}

\bibliographystyleSM{splncs04}
\bibliographySM{refs}
\end{document}

%% file: mathstuff.tex
\newcommand{\vect}[1]{\mathbf{ #1}}
\newcommand{\vectg}[1]{{\boldsymbol{ #1}}}
\newcommand{\ggo}{\ensuremath{\mathrm{g^2o}} }
\newcommand{\R}{\mathbb{R}}
\newcommand{\N}{\mathbb{N}}
\newcommand{\Z}{\mathbb{Z}}
\renewcommand{\P}{\mathbb{P}}
\newcommand{\tran}{^\top}
\newcommand{\T}{^\mathsf{T}}
\newcommand{\iT}{^{-\mathsf{T}}}
\newcommand{\inv}{^{-1}}
\newcommand{\func}[2]{\mathtt{#1}\left\{#2\right\}}
\newcommand{\sig}{\operatorname{sig}}
\newcommand{\diag}{\operatorname{diag}}
\newcommand{\argmin}{\operatornamewithlimits{argmin}}
\newcommand{\argmax}{\operatornamewithlimits{argmax}}
\newcommand{\RMSE}{\operatorname{RMSE}}
\newcommand{\RMSEpos}{\operatorname{RMSE}_\text{pos}}
\newcommand{\RMSEori}{\operatorname{RMSE}_\text{ori}}
\newcommand{\RPE}{\operatorname{RPE}}
\newcommand{\RPEpos}{\operatorname{RPE}_\text{pos}}
\newcommand{\RPEori}{\operatorname{RPE}_\text{ori}}
\newcommand{\rpe}{\varepsilon_{\vdelta}}
\newcommand{\achiError}{\bar{e}_{\chi^2}}
\newcommand{\chiError}{e_{\chi^2}}
\newcommand{\normal}[2]{\mathcal{N}\left(#1, #2\right)}
\newcommand{\uniform}[2]{\mathcal{U}\left(#1, #2\right)}
\newcommand{\pfrac}[2]{\frac{\partial #1}{\partial #2}}  % partielles Differential
\newcommand{\fracpd}[2]{\frac{\partial #1}{\partial #2}} % partielles Differential
\newcommand{\fracppd}[2]{\frac{\partial^2 #1}{\partial #2^2}}  % zweite partielle 
\newcommand{\dd}{\mathrm{d}}  
\newcommand{\smd}[2]{\left\| #1 \right\|^2_{#2}}
\newcommand{\E}[1]{\text{\normalfont{E}}\left[ #1 \right]}     % Expectation operator
\newcommand{\Cov}[1]{\text{\normalfont{Cov}}\left[ #1 \right]} % Covariance operator
\newcommand{\Var}[1]{\text{\normalfont{Var}}\left[ #1 \right]} % Variance operator
\newcommand{\Tr}[1]{\text{\normalfont{tr}}\left( #1 \right)}   % Trace of a matrix
\def\sgn{\mathop{\mathrm sgn}}    
\newcommand{\twovector}[2]{\begin{pmatrix} #1 \\ #2 \end{pmatrix}} % Vektor mit zwei zwei Einträgen
\newcommand{\smalltwovector}[2]{\left(\begin{smallmatrix} #1 \\ #2 \end{smallmatrix}\right)} 
\newcommand{\threevector}[3]{\begin{pmatrix} #1 \\ #2 \\ #3 \end{pmatrix}} % Vektor mit drei Einträgen
\newcommand{\fourvector}[4]{\begin{pmatrix} #1 \\ #2 \\ #3 \\ #4 \end{pmatrix}}  % Vektor mit vier Einträgen
\newcommand{\smallthreevector}[3]{\left(\begin{smallmatrix} #1 \\ #2 \\ #3 \end{smallmatrix}\right)} % kleiner Vektor mit drei Einträgen
\newcommand{\fourmatrix}[4]{\begin{pmatrix} #1 & #2 \\ #3 & #4 \end{pmatrix}} % 2x2-Matrix
\newcommand{\vA}{\vect{A}}
\newcommand{\vB}{\vect{B}}
\newcommand{\vC}{\vect{C}}
\newcommand{\vD}{\vect{D}}
\newcommand{\vE}{\vect{E}}
\newcommand{\vF}{\vect{F}}
\newcommand{\vG}{\vect{G}}
\newcommand{\vH}{\vect{H}}
\newcommand{\vI}{\vect{I}}
\newcommand{\vJ}{\vect{J}}
\newcommand{\vK}{\vect{K}}
\newcommand{\vL}{\vect{L}}
\newcommand{\vM}{\vect{M}}
\newcommand{\vN}{\vect{N}}
\newcommand{\vO}{\vect{O}}
\newcommand{\vP}{\vect{P}}
\newcommand{\vQ}{\vect{Q}}
\newcommand{\vR}{\vect{R}}
\newcommand{\vS}{\vect{S}}
\newcommand{\vT}{\vect{T}}
\newcommand{\vU}{\vect{U}}
\newcommand{\vV}{\vect{V}}
\newcommand{\vW}{\vect{W}}
\newcommand{\vX}{\vect{X}}
\newcommand{\vY}{\vect{Y}}
\newcommand{\vZ}{\vect{Z}}
\newcommand{\va}{\vect{a}}
\newcommand{\vb}{\vect{b}}
\newcommand{\vc}{\vect{c}}
\newcommand{\vd}{\vect{d}}
\newcommand{\ve}{\vect{e}}
\newcommand{\vf}{\vect{f}}
\newcommand{\vg}{\vect{g}}
\newcommand{\vh}{\vect{h}}
\newcommand{\vi}{\vect{i}}
\newcommand{\vj}{\vect{j}}
\newcommand{\vk}{\vect{k}}
\newcommand{\vl}{\vect{l}}
\newcommand{\vm}{\vect{m}}
\newcommand{\vn}{\vect{n}}
\newcommand{\vo}{\vect{o}}
\newcommand{\vp}{\vect{p}}
\newcommand{\vq}{\vect{q}}
\newcommand{\vr}{\vect{r}}
\newcommand{\vt}{\vect{t}}
\newcommand{\vu}{\vect{u}}
\newcommand{\vv}{\vect{v}}
\newcommand{\vw}{\vect{w}}
\newcommand{\vx}{\vect{x}}
\newcommand{\vy}{\vect{y}}
\newcommand{\vz}{\vect{z}}
\newcommand{\valpha}{\vectg{\alpha}}
\newcommand{\vbeta}{\vectg{\beta}}
\newcommand{\vgamma}{\vectg{\gamma}}
\newcommand{\vdelta}{\vectg{\delta}}
\newcommand{\vepsilon}{\vectg{\epsilon}}
\newcommand{\vtau}{\vectg{\tau}}
\newcommand{\vmu}{\vectg{\mu}}
\newcommand{\vphi}{\vectg{\phi}}
\newcommand{\vPhi}{\vectg{\Phi}}
\newcommand{\vpi}{\vectg{\pi}}
\newcommand{\vPi}{\vectg{\Pi}}
\newcommand{\vPsi}{\vectg{\Psi}}
\newcommand{\vchi}{\vectg{\chi}}
\newcommand{\vvarphi}{\vectg{\varphi}}
\newcommand{\veta}{\vectg{\eta}}
\newcommand{\viota}{\vectg{\iota}}
\newcommand{\vkappa}{\vectg{\kappa}}
\newcommand{\vlambda}{\vectg{\lambda}}
\newcommand{\vLambda}{\vectg{\Lambda}}
\newcommand{\vnu}{\vectg{\nu}}
\newcommand{\vgo}{\vectg{\o}}
\newcommand{\vvarpi}{\vectg{\varpi}}
\newcommand{\vtheta}{\vectg{\theta}}
\newcommand{\vTheta}{\vectg{\Theta}}
\newcommand{\vvartheta}{\vectg{\vartheta}}
\newcommand{\vrho}{\vectg{\rho}}
\newcommand{\vsigma}{\vectg{\sigma}}
\newcommand{\vSigma}{\vectg{\Sigma}}
\newcommand{\vvarsigma}{\vectg{\varsigma}}
\newcommand{\vupsilon}{\vectg{\upsilon}}
\newcommand{\vomega}{\vectg{\omega}}
\newcommand{\vOmega}{\vectg{\Omega}}
\newcommand{\vxi}{\vectg{\xi}}
\newcommand{\vXi}{\vectg{\Xi}}
\newcommand{\vpsi}{\vectg{\psi}}
\newcommand{\vzeta}{\vectg{\zeta}}
\newcommand{\vzero}{\vect{0}}
\newcommand{\cA}{\mathcal{A}}
\newcommand{\cB}{\mathcal{B}}
\newcommand{\cC}{\mathcal{C}}
\newcommand{\cD}{\mathcal{D}}
\newcommand{\cE}{\mathcal{E}}
\newcommand{\cF}{\mathcal{F}}
\newcommand{\cG}{\mathcal{G}}
\newcommand{\cH}{\mathcal{H}}
\newcommand{\cI}{\mathcal{I}}
\newcommand{\cJ}{\mathcal{J}}
\newcommand{\cK}{\mathcal{K}}
\newcommand{\cL}{\mathcal{L}}
\newcommand{\cM}{\mathcal{M}}
\newcommand{\cN}{\mathcal{N}}
\newcommand{\cO}{\mathcal{O}}
\newcommand{\cP}{\mathcal{P}}
\newcommand{\cQ}{\mathcal{Q}}
\newcommand{\cR}{\mathcal{R}}
\newcommand{\cS}{\mathcal{S}}
\newcommand{\cT}{\mathcal{T}}
\newcommand{\cU}{\mathcal{U}}
\newcommand{\cV}{\mathcal{V}}
\newcommand{\cW}{\mathcal{W}}
\newcommand{\cX}{\mathcal{X}}
\newcommand{\cY}{\mathcal{Y}}
\newcommand{\cZ}{\mathcal{Z}}
\newcommand{\fA}{\mathfrak{A}}
\newcommand{\fB}{\mathfrak{B}}
\newcommand{\fC}{\mathfrak{C}}
\newcommand{\fD}{\mathfrak{D}}
\newcommand{\fE}{\mathfrak{E}}
\newcommand{\fF}{\mathfrak{F}}
\newcommand{\fG}{\mathfrak{G}}
\newcommand{\fH}{\mathfrak{H}}
\newcommand{\fI}{\mathfrak{I}}
\newcommand{\fJ}{\mathfrak{J}}
\newcommand{\fK}{\mathfrak{K}}
\newcommand{\fL}{\mathfrak{L}}
\newcommand{\fM}{\mathfrak{M}}
\newcommand{\fN}{\mathfrak{N}}
\newcommand{\fO}{\mathfrak{O}}
\newcommand{\fP}{\mathfrak{P}}
\newcommand{\fQ}{\mathfrak{Q}}
\newcommand{\fR}{\mathfrak{R}}
\newcommand{\fS}{\mathfrak{S}}
\newcommand{\fT}{\mathfrak{T}}
\newcommand{\fU}{\mathfrak{U}}
\newcommand{\fV}{\mathfrak{V}}
\newcommand{\fW}{\mathfrak{W}}
\newcommand{\fX}{\mathfrak{X}}
\newcommand{\fY}{\mathfrak{Y}}
\newcommand{\fZ}{\mathfrak{Z}}
\newcommand{\fc}{\mathfrak{c}}
\newcommand{\fo}{\mathfrak{o}}

%% file: sections/introduction.tex
\section{Introduction}

In 2013, Scheirer et al.~\cite{scheirer2013towards} described the closed-set assumption ingrained in almost all vision models at the time: all test classes are known during training and the model is never exposed to novel or unknown classes during testing. Real-world applications challenge this assumption and test in open-set conditions -- encountering unexpected objects not in the training dataset -- and causing potentially dangerous overconfident misclassifications~\cite{dhamija2020overlooked,sunderhauf2018limits, miller2018dropout, bendale2016towards}. Open-set recognition~\cite{scheirer2013towards} was introduced to address this issue by evaluating the ability of vision models to identify and reject open-set inputs as unknown.

Vision-language models (VLMs) have recently revolutionised the fields of image classification~\cite{radford2021clip, jia2021align, girdhar2023imagebind} and object detection~\cite{ovdetr,ovrcnn,vild}. Trained on vast and diverse internet-scale datasets, VLMs recognise an extensive variety of classes and \emph{seemingly} are open-set by default~\cite{wu2024towards}. Our paper challenges this view.

\begin{figure}[tb]
    \centering
    \includegraphics[width=1\linewidth]{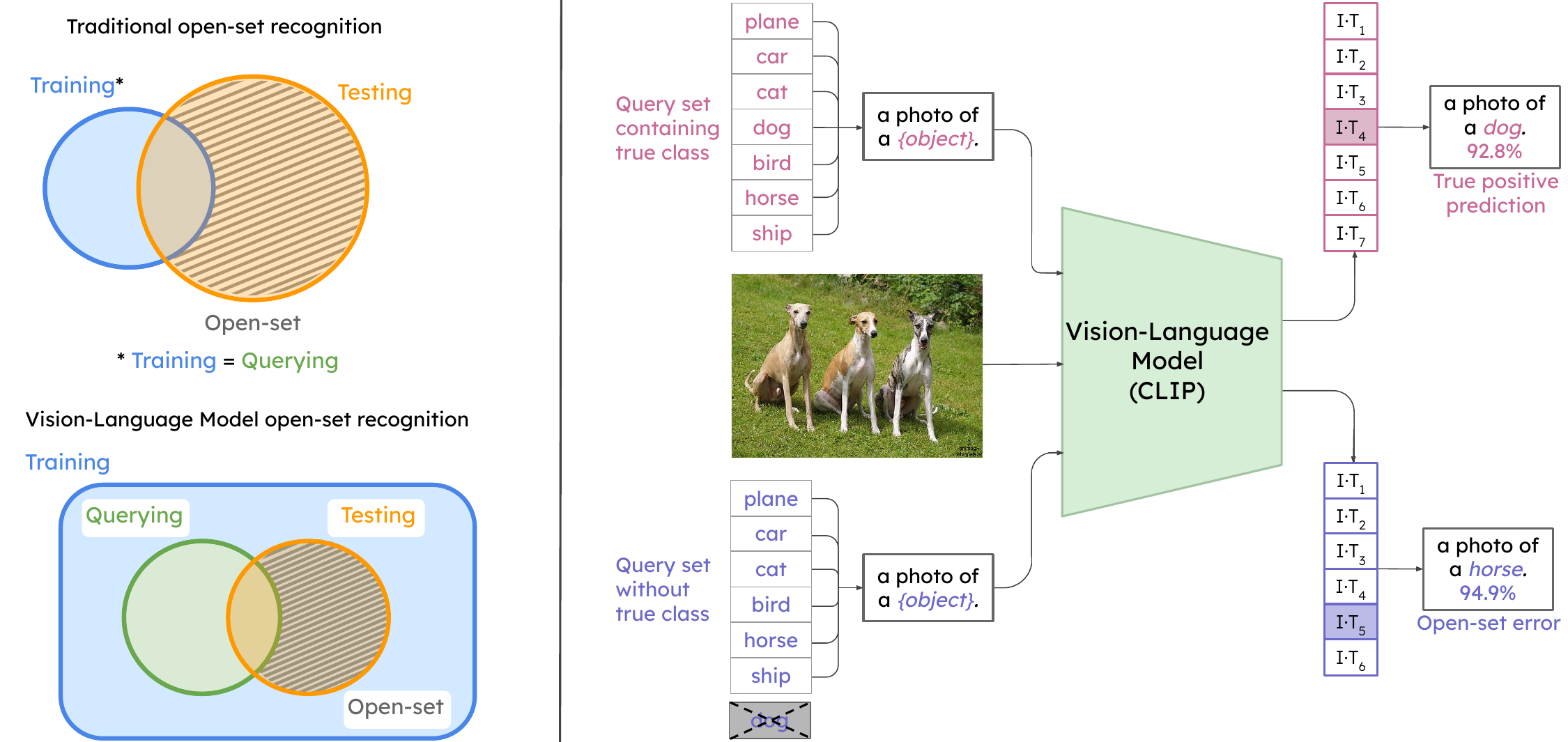}
    \caption{Left: Traditional open-set recognition arises when testing classes outside a model's finite training class set. While VLMs are trained on internet-scale datasets containing most conceivable object classes, they use a finite query set for classification. Open-set recognition arises when test classes are not included in the query set. Right: When testing an object present in the predefined query set, VLMs often correctly classify the object. When testing an object that is not present in the query set (i.e. an open-set object), VLMs often misclassify the object as a query class with high confidence (i.e. an open-set error).}
    \label{fig:vlmopenset}
\end{figure}

In the age of VLMs, we introduce an updated definition of the open-set problem -- while previous definitions focused on a finite \emph{training} set~\cite{scheirer2013towards}, we show that VLMs impose closed-set assumptions through a finite \emph{query} set. 
As illustrated in Fig.~\ref{fig:vlmopenset}, VLMs for open-vocabulary image classification or object detection are still evaluated under closed-set conditions: they compare image embeddings with text embeddings from a predefined, finite query set of class labels. This introduces a closed-set assumption that all classes encountered during testing are included in this query set. Open-set conditions emerge when VLMs encounter objects that are not included in the query set (unknown objects). As we show in Section~\ref{sec:experiments}, even state-of-the-art VLMs heavily degrade in performance and misclassify unknown objects as belonging to the query set with high confidence. 

In theory, this problem disappears if a VLM's query set contains \emph{every possible class label in a vocabulary}. However, we show that naively increasing the query set to contain more and more class labels results in worse task performance, where increasing numbers of misclassifications occur (see Sec.~\ref{sec:querysize}). The serious concerns about deploying vision models in safety-critical applications voiced in the pre-VLM era~\cite{sunderhauf2018limits, amodei2016concrete} continues to be relevant and underscore the urgent need for further research into the open-set problem for VLMs.

Our paper is the first to systematically investigate and quantify the open-set vulnerabilities of VLM-based open-vocabulary object detectors and zero-shot image classifiers. We re-define the open-set problem for VLMs (Section~\ref{sec:problem}) and compare promising baseline strategies,  leveraging predictive uncertainty (Section~\ref{sec:stateoftheart}) and different approaches to inserting dedicated \texttt{negative} (unknown) class representations into the VLM pipeline in Section~\ref{sec:negative}. We propose a new benchmark and evaluation protocol in Section~\ref{sec:protocol} to foster further research and evaluation of VLMs for open-set recognition.

%% file: sections/background.tex
\section{Background}

\subsection{Vision-Language Models for Zero-shot Classification}
% Recent years have seen the emergence of foundation models for computer vision, one such type being the \emph{vision-language model (VLM)}.
% Trained on internet-scale datasets with self-supervision learning techniques, foundation models can be applied to a range of downstream tasks \cite{bommasani2021opportunities}.
A recently emerged foundation model is the \emph{vision-language model (VLM)}, which learns a multi-modal feature space that can jointly represent visual features and text~\cite{radford2021clip, jia2021align, yu2022coca, zhai2022lit, alayrac2022flamingo} (and in some cases other modalities as well~\cite{yuan2021florence, girdhar2023imagebind}). 
% VLMs are typically trained with image-text pairs scraped from the Internet, i.e. online images and their corresponding captions, applying contrastive learning to maximise the similarity between images and their respective text \cite{radford2021clip, jia2021align}. These datasets can be in the magnitude of hundreds of millions to billions of data pairs \cite{radford2021clip, jia2021align, yuan2021florence}. 
Large-scale contrastive, self-supervised learning on internet-scale datasets provides VLMs with a rich feature representation that can be adapted for many downstream tasks, one of which is zero-shot or open vocabulary image classification~\cite{radford2021clip, jia2021align, yu2022coca, zhai2022lit, alayrac2022flamingo, yuan2021florence, girdhar2023imagebind}. 
% To perform classification, an image and set of query classes (text) are passed through the VLM and compared in the joint feature space, where the most similar class is predicted for classification~\cite{radford2021clip}. 
Despite not being trained for classification tasks or datasets, VLMs generalise surprisingly well and achieve near state-of-the-art supervised learning performance~\cite{yu2022coca}.
Alongside this, a distinguishing advantage of VLMs is their adaptability to different datasets with different classes -- the set of classes can be changed at will depending on the dataset~\cite{radford2021clip, jia2021align}.

\subsection{Vision-Language Models for Open Vocabulary Object Detection}
Unlocked by the advances in VLMs for zero-shot classification,~\cite{ovrcnn} proposed open vocabulary object detection, where detectors must generalise to an arbitrary set of object classes at test time, even object classes that were not seen during training~\cite{ovrcnn}. This is related to our proposed open-set recognition, but assumes that novel object classes are added to the query set, whereas we test against object classes that are not present in the query set.

Existing open vocabulary detectors leverage VLM backbones to enable generalisation to new object classes, embedded within Faster R-CNN-style architectures~\cite{ovrcnn, vild, detic, regionclip, vldet, detpro, fvlm, baron, promptdet, lpovod, goat} or DETR-style architectures~\cite{ovdetr, cora}. CLIP~\cite{radford2021clip} is frequently utilised by these open vocabulary detectors, directly included in the architecture via its text encoder~\cite{regionclip, ovdetr, vild, vldet, detic, detpro, baron, promptdet, lpovod, goat} or in its entirety~\cite{cora, fvlm}, or used as a supervision signal during training ~\cite{regionclip, vild, ovdetr, detpro, baron, promptdet, lpovod, goat}. 
We refer the reader to \cite{wu2024towards} for a comprehensive survey on open vocabulary learning.

\subsection{Open-set Recognition}
Scheirer et al.~\cite{scheirer2013towards} introduced open-set recognition to challenge the prevailing closed-set assumptions of image recognition systems. \emph{Closed-set} models only test on samples from a predefined set of classes that were known during model training~\cite{scheirer2013towards}. In \emph{open-set} recognition, samples can instead come from an expansive range of classes that cannot be predefined, and therefore are unknown during model training~\cite{scheirer2013towards}.Open-set recognition tests the ability of models to identify and reject unknown samples that do not belong to one of the training classes. The open-set concept has been explored widely in computer vision, including in image classification~\cite{bendale2016towards, vaze2022open, zhang2020hybrid, yoshihashi2019classification, oza2019c2ae, perera2020generative, chen2021adversarial}, object detection~\cite{dhamija2020overlooked, miller2018dropout, han2022expanding, miller2022openset, zhou2023open}, image segmentation~\cite{hwang2021exemplar, pham2018bayesian, cen2021deep} and other vision tasks~\cite{panareda2017open, saito2018open, maalouf2024follow, bao2021evidential, liu2022open}. 

Our paper is the first to systematically investigate the open-set problem for VLMs, and our findings challenge the assumption that the open-set problem has been solved for VLMs due to their exposure to internet-scale training data. The following sections define the closed-set assumptions existing in current VLMs, demonstrate that VLMs continue to be vulnerable to open-set conditions and evaluate a number of baseline strategies for open-set recognition with VLMs.

%% file: sections/problem.tex
\renewcommand{\vs}{\vect{s}}

\section{Problem Definition}

\label{sec:problem}
The core concept behind VLMs, mapping images and text into a joint embedding space, makes them powerful foundations for image classification and object detection. Classification is performed by comparing an image embedding $\ve^I$ with text embeddings $\cE^q = \{\ve_1^q, \dots, \ve^q_{N}\}$ obtained from a set of query labels $\cQ = \{q_1, \dots, q_{N} \}$ which are class names such as ``\texttt{dog}''\footnote{Most methods insert an additional step here and generate a set of prompts from each query label $q_i$, such as \{``A photo of a dog'', ``A picture of a dog'', \dots\} and compare the embeddings of these prompts with $\ve^I$. For clarity of notation, we omit this step.}. The similarities between the image embedding and all text embeddings from $\cQ$ are measured by a similarity function $\vs^q = \operatorname{sim}(\ve^I, \cE^q)$, e.g. using the cosine similarity.
The text embedding with the highest similarity to the image embedding, i.e. $\argmax \vs^q$, determines the predicted class label. The process is the same for object detection, with $\ve^I$ now representing the embedding of an image region corresponding to an object proposal.

\noindent\textbf{Open-set and Closed-set:} While the pre-VLM definition of the closed-set and open-set concepts focused on the challenges arising from a finite \emph{training} set~\cite{scheirer2013towards}, our updated definition emphasises the limitations of the finite \emph{query} set: 
The closed-set assumption in VLMs is that all classes seen during testing are predefined prior to testing and included in the query set. 
An open-set situation arises when the VLM encounters an object or image class that is not part of this predefined query set $\cQ$.

\subsection{Baseline Approaches to Open-Set Recognition for VLMs}
\label{sec:baselineapproaches}
In an open-set situation, the VLM should be able to identify an input not represented by the query set and reject it is as \texttt{unknown}, instead of misclassifying it as one of the query classes from $\cQ$. Similar to the pre-VLM open-set literature, there are multiple ways of recognising an input as unknown \cite{geng2020recent}. We investigate baseline methods using predictive uncertainty and different approaches of negative embeddings.

\noindent\textbf{Using Uncertainty:}
Different measures of predictive uncertainty $\psi$ can be used to reject inputs as \texttt{unknown}, for example based on the class similarities, e.g. $\psi^\text{Cosine}=\max(\vs^q)$ or $\psi^\text{Softmax}=\max(\operatorname{softmax}(\vs^q))$, or the entropy of the softmax similarities $\psi^\text{Entropy}=-H(\operatorname{softmax}(\vs^q))$. 
Only inputs that pass a threshold test $\psi\ge\theta$ are classified according to $\argmax \vs^q$. Otherwise, the input is declared to be open-set and unknown.

\noindent\textbf{Dedicated Negative Embeddings:}
Another approach is to create a set of augmented query embeddings $\tilde{\cE}^q = \cE^q \cup \cE^\texttt{n}$ by adding $M$ dedicated \texttt{negative} class embeddings $\cE^\texttt{n} = \{\ve^\texttt{n}_1, \dots, \ve^\texttt{n}_{M} \}$. The input image or detection can then be rejected as \texttt{unknown} if the image embedding is most similar to one of those \texttt{negative} embeddings $\ve^\texttt{n}_i$ rather than a query class embedding $\ve^q_j$.  These negative embeddings can be created in one of two ways: (1) by creating \texttt{negative words}, which are then passed through the VLM text encoder to create the negative embeddings, or (2) by directly creating \texttt{negative embeddings}. 

We investigate two baseline options for negative embeddings: 
a \textbf{random words} method, where we generate random strings with random length between 2-8 characters (e.g. ``braxl''); 
and a \textbf{random embeddings} method, where we randomly draw $M$ embedding vectors from a Gaussian distribution, i.e. $\ve^\texttt{n}_i \sim \mathcal{N}(\vmu, \vSigma)$, using the mean and standard deviation from the query embedding vectors $\cE^q$.
We compare the effectiveness of these approaches, including for different numbers of $M$ random negative queries, in our experiments. 
Interestingly, many open vocabulary object detectors already use a negative query approach, though typically with a only single embedding vector of all zeroes~\cite{ovrcnn, regionclip, detic, vldet, cora}.

%% file: sections/evaluate.tex
\section{Evaluation Protocol}
\label{sec:protocol}
\subsection{Creating an Open-set Recognition Dataset for VLMs}
\label{sec:protocol_data}
We now introduce how to use existing image classification or object detection datasets to evaluate open-set recognition. In short, we test all images twice: once in a standard closed-set manner with all dataset classes included in the query set (see Fig.~\ref{fig:vlmopenset} top-right), and once in an open-set manner with a query set containing all dataset classes not present in the image (see Fig.~\ref{fig:vlmopenset} bottom-right). 

Classification and object detection datasets have a pre-defined set of $K$ labelled target classes $\cL$.
In image classification, there is a single corresponding class label $y \in \cL$ per image. In object detection, there are multiple ground truth objects in each image and each has a corresponding class label. For both tasks, we can define a set $\cY$ that contains the ground truth class labels for each image: $\cY$ contains a single label in classification and multiple labels in object detection. 

\noindent\textbf{Testing Closed-set Recognition: } We first test images in a standard manner to establish the closed-set performance of the VLM and identify all \emph{true positive (TP)} predictions. This involves testing all images with a query set identical to the dataset class list, $\cQ = \cL$. 
To identify the TP predictions, we follow standard protocol for classification and object detection datasets -- in classification, a TP has a predicted class matching the ground-truth label; in object detection, this involves a object-detection assignment process that requires correct object classification and localisation (see Supplementary Sec. S1). TPs represent correct closed-set predictions that should not be rejected as \texttt{unknown}, and we collect the uncertainty measures $\psi$ associated with all TPs for later evaluation.

For object detection datasets, unlabelled objects are present in the background of the image~\cite{dhamija2020overlooked}. These unlabelled objects are not in the query set $\cQ$, and therefore meet our definition for open-set objects. Detections of these open-set objects will present as false positives (FPs) in a closed-set test, however not all FPs arise from open-set objects -- FPs in object detection can also arise due to poor object localisation, misclassification between query classes, or duplicate detections of a query class (see~\cite{bolya2020tide}). Rather than attempting to categorise different FP types to distinguish open-set errors, our approach definitively identifies open-set errors for both object detection and image classification.

\noindent\textbf{Testing Open-set Recognition: } 
To evaluate open-set recognition, we test each image in the dataset 
% Each image in the dataset is tested with a 
with a modified query set, $\tilde{\cQ}$ that contains all the class labels \underline{not} in the image ground-truth class list, $\tilde{\cQ} = \cL - \cY$. This ensures the image only contains object classes that are not in the modified query set -- every single prediction must be an \emph{open-set error (OSE)}\footnote{This assumes accurate labelling of ground-truth classes, though most datasets will contain some label error. This approach is unsuitable for weakly-labelled datasets.}. In image classification, this modifies the query set to have $K-1$ labels. In object detection, the modified query set size depends on the number of unique object classes in the image. We collect the uncertainty measures $\psi$ of all OSE predictions for evaluation.

\subsection{Metrics}
We use performance metrics that measure the ability of the VLM to reject OSEs (with high uncertainty or with a \texttt{negative} class), while maintaining the closed-set task performance. 

\noindent\textbf{Precision-Recall Curves: } Open-set recognition can be formulated as a binary classification task that uses uncertainty as the decision threshold. Correct closed-set predictions (introduced as TPs in Sec.~\ref{sec:protocol_data}) and open-set errors (OSEs) are considered as the positive and negative class respectively. We construct precision-recall curves, which are well-suited to this problem as they are affected by the size of the negative class (OSEs)~\cite{saito2015precision}\footnote{ROC curves do not have this property -- we discuss this further in Sec. S3 of the Supplementary Material where we report AuROC for the object detection experiments.} as well as prediction uncertainty. High precision indicates that little-to-none negatives (OSEs) remain after the uncertainty thresholding. High recall indicates that most positives (TPs) remain after the uncertainty thresholding. We summarise the performance of the precision-recall curves with 3 core metrics: (1) \textbf{Area under the PR curve (AuPR)}, (2) \textbf{Precision at $\mathbf{95\%}$ Recall}, and (3) \textbf{Recall at $\mathbf{95\%}$ Precision}.  

\noindent\textbf{Task Metrics: } We use the established task metrics used to evaluate classification and detection -- top-1 accuracy~\cite{russakovsky2015imagenet} and mean Average Precision (mAP)~\cite{lin2014microsoft} respectively. We evaluate mAP at an IoU threshold of $0.5$ -- we are primarily interested in the classification ability of the detector (rather than localisation) and this is standard protocol in open vocabulary detection~\cite{ovrcnn, ovdetr, vldet, cora, vild, detic, regionclip}.

\noindent\textbf{Absolute Counts:} We also report the absolute counts of TPs and OSEs. This can be particularly useful for object detection, where increased task performance does not necessarily indicate increased TPs and detectors predict variable numbers of OSEs. We stress that \emph{absolute counts should not be considered alone as a performance indicator}, e.g. an increased number of OSEs can be acceptable  if all OSEs have very high uncertainty. 

\subsection{VLM Baselines}
We test with six VLM classifiers -- CLIP~\cite{radford2021clip}, ALIGN~\cite{jia2021align}, CoCa~\cite{yu2022coca}, ImageBind~\cite{girdhar2023imagebind}, SigLIP~\cite{siglip} and LanguageBind~\cite{languagebind} -- and seven open vocabulary object detectors -- OVR-CNN~\cite{ovrcnn}, ViLD~\cite{vild}, OV-DETR~\cite{ovdetr}, RegionCLIP~\cite{regionclip}, Detic~\cite{detic}, VLDet~\cite{vldet} and CORA~\cite{cora} (see Supplementary Sec. S2 for implementation details). These VLMs cover a range of architectures and training paradigms and have publicly-available implementations. Our goal with testing this range of VLMs is not to find the ``best'' VLM for open-set recognition, but instead to exhibit the general vulnerability of VLMs to open-set conditions. 
For each VLM, we extract the prediction \emph{Softmax} score, \emph{Cosine} similarity, and \emph{Entropy} of the softmax distribution as baseline uncertainty measures. Some VLMs use Sigmoid activation rather than Softmax~\cite{siglip, ovdetr, detic, vldet, cora} -- for these methods, we only report the \emph{Sigmoid} score as its results are identical to the cosine similarity and the entropy is unsuitable as the scores are not normalised to a distribution. 
We do not impose a minimum score threshold on any predictions, as the AuPR metric captures performance over all possible thresholds.

\subsection{Datasets}
We focus on general object recognition and test with the large-scale and prevailing datasets used to benchmark image classification and object detection: For image classification, we test with the ImageNet1k validation dataset~\cite{deng2009imagenet, russakovsky2015imagenet}. It contains $50,000$ images from $1000$ classes and is \href{https://www.image-net.org/download.php}{publicly available} for research.
For object detection, we test with the MS COCO 2017 validation dataset~\cite{lin2014microsoft}. It contains $4,952$ images with $36,781$ annotated objects from $80$ classes and is \href{https://cocodataset.org/#download}{publicly available} for research. We show results for domain-specific classification datasets in the Supplementary Sec. S4.

%% file: sections/experiments.tex
\section{Experiments and Results}
\label{sec:experiments}

\subsection{State-of-the-art VLMs Perform Poorly in Open-set Conditions.}
\label{sec:stateoftheart}

\noindent \textbf{Classification: } 
\input{tables/classification_imagenet_uncertainty}
Tab.~\ref{tab:cls_unc} shows that all six VLM classifiers perform poorly for open-set recognition. For scenarios that require high recall of true positive predictions (i.e. $95\%$ recall), approximately every second prediction returned is an open-set error (i.e. precision between $46.2\%$ and $58.4\%$). For scenarios that require high precision of predictions (i.e. $95\%$ precision), less than 12\% of the true positive predictions are retained (i.e. recall between 3.4\% and 11.1\%).
Comparing across uncertainty types, the VLM classifiers show better uncertainty performance when using the predicted Softmax score or Entropy rather than the predicted Cosine similarity. This suggests raw similarity in VLM feature space is not the best indicator of open-set error, with measures of relative similarity comparative to other classes offering better performance. 

\input{tables/detection_coco}
\noindent \textbf{Object Detection: } 
All seven of the tested VLM object detectors are vulnerable to open-set errors -- see Tab.~\ref{tab:det_unc}. Unlike VLM classifiers, most VLM detectors already contain a negative query to capture objects not in the query set (often referred to as the \emph{background} class). Despite this, the VLM detectors produce open-set error counts in the range of $100,000$ to $1,500,000$ when tested on only 4,952 images, i.e. between $20$ and $300$ open-set error per image on average.

High open-set error counts are not inherently a problem if they are produced with high uncertainty. Yet for all tested detectors, the baseline uncertainty methods cannot adequately distinguish true positive detections from open-set errors. When thresholding for high recall of true positives, \emph{at best} only every fifth detection is a true positive (i.e. precision ranges between $5.9\%$ and $21.9\%$). When thresholding for high precision, \emph{at best} less than half of true positives are recalled (i.e. recall ranges between 16.3\% and 48.5\%). We include some qualitative examples of highly confident open-set errors in the Supplementary Sec. S6.

Notably, the precision-recall characteristics for VLM classification and object detection differ significantly -- VLM classifiers achieve better performance at high recall, whereas VLM detectors achieve better performance at high precision. This suggest that while the VLM classifiers struggle with over-confident open-set errors, the VLM detectors suffer from under-confident true positives. We include histograms that visualise this trend in the Supplementary Sec. S7.

We also consider how the performance of uncertainty differs for identifying open-set error versus closed-set misclassifications, with results and a detailed discussion in the Supplementary Sec. S5. We find that the performance of softmax uncertainty is lower for open-set error than for closed-set error in both classification and detection tasks. 
In classification, open-set performance is particularly decremented when thresholding for high precision -- likely due to a greater proportion of overconfident open-set error -- whereas detection primarily degrades with increased numbers of errors, with ViLD for example producing
$11.7\times$ more open-set than closed-set misclassifications.  

\subsection{Can Negative Queries Improve Open-set Performance?}
\label{sec:negative}

\input{tables/classification_imagenet_other}
\begin{figure}[t!]
    \centering
\includegraphics[width=\linewidth]{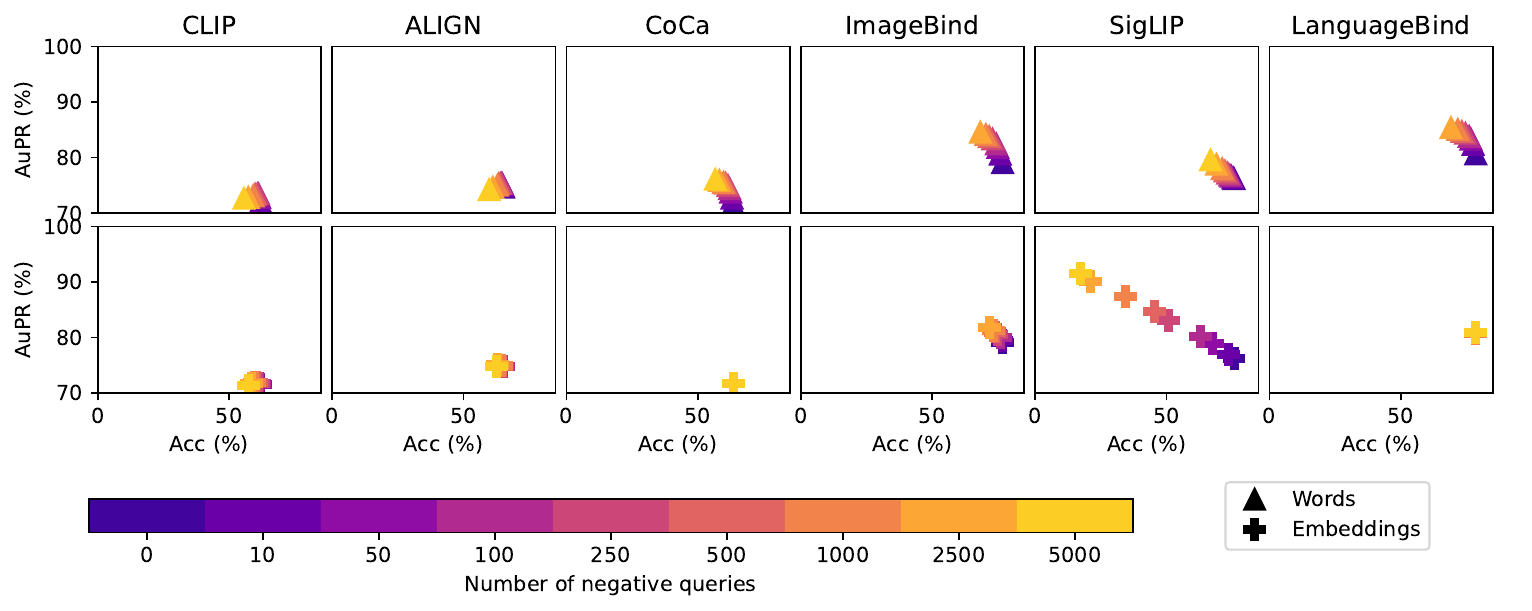}
    \caption{The trade-off between closed-set and open-set performance with increasing number of negative queries for different VLM classifiers.}
    \label{fig:cls_negvsperf}
\end{figure}

\noindent \textbf{Classification: } 
Negative class queries provide the VLM classifiers with the ability to select ``unknown'' when testing an input. When used in larger quantities, negative queries can capture and remove significant portions of open-set error, yet this happens in trade-off with a reduction in closed-set performance -- see Tab.~\ref{tab:cls_other}. For example, ImageBing using $2,500$ random word negative queries effectively halves the open-set error, but this occurs in trade-off with a closed-set accuracy reduction of $8.4\%$ (approximately $4,000$ of the closed-set images are incorrectly classified as one of the random words). 

While negative queries noticeably reduce the raw open-set error count, this is not always mirrored with increased performance in the open-set uncertainty metrics. We present the change in closed-set and open-set performance against the number of negative query points in Fig. \ref{fig:cls_negvsperf}. Generally the use of negative random words appears the most effective, with CoCa, ImageBind, SigLIP and LanguageBind showing improved open-set performance with minimal loss of closed-set performance. For some of the classifiers, there is negligible change in open-set performance despite the reduction in open-set error. We explore this further in the Supplementary Sec. S8, where we show that in some VLMs the negative queries only capture and remove ``easy'' open-set errors, i.e. open-set errors that already have high uncertainty.

\begin{figure}[t]
    \centering
\includegraphics[width=\linewidth]{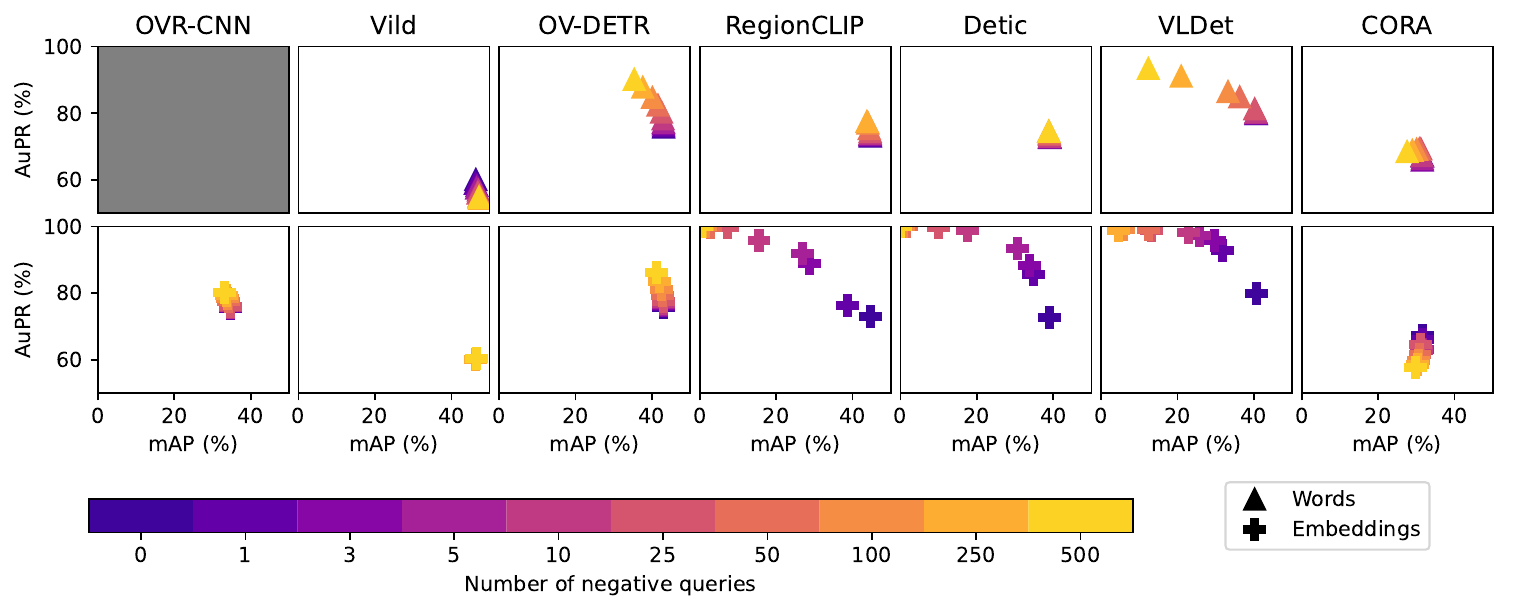}
    \caption{The trade-off between closed-set and open-set performance with increasing number of negative queries for different VLM object detectors.}
    \label{fig:det_negvsperf}
\end{figure}

\noindent \textbf{Object Detection: } 
The VLM detectors do not show a consistent response to the use of negative queries. Different detectors show different sensitivities to negative query counts, different trade-off rates between closed-set and open-set performance, and differences in effectiveness of embedding versus word negative query types -- see Tab.~\ref{tab:det_other} and Fig. \ref{fig:det_negvsperf}. Looking at Fig.~\ref{fig:det_negvsperf}, we can consider the sensitivity, trade-off rate and response of the detectors to negative queries: greater performance changes with increasing queries indicates sensitivity; large decreases in mAP with more queries indicates increased trade-off rates between open-set and closed-set performance; and a positive response to the negative queries presents as an increase in AuPR when increasing queries (negative response decreases AuPR).

\input{tables/detection_coco_other}

OV-DETR is the only detector to show a clear positive response, low trade-off rate, and high sensitivity to the negative query approach, particularly with negative embeddings. While OVR-CNN (embeddings), RegionCLIP (words), Detic (words) also show a positive response and low trade-off rate, this is with very little sensitivity to increasing numbers of negative queries. In contrast, the open-set performance of Detic, RegionCLIP, and VLDet are very sensitive to negative embeddings, but at a rapid trade-off with closed-set performance. Both ViLD and CORA show a negative response or no sensitivity to negative queries.

It is tempting to relate these trends to the architecture or training paradigm of the VLM detectors, however we could not identify many common characteristics. RegionCLIP, Detic and VLDet all learn from auxiliary datasets that have been pseudo-labelled with regions~\cite{regionclip, detic, vldet}, although this does not clearly indicate a reason for high sensitivity and trade-off rate for negative embeddings. Vild~\cite{vild} and CORA~\cite{cora} both heavily rely on CLIP's visual feature encoding (either by using CLIP directly~\cite{cora} or knowledge distillation~\cite{vild}) -- their poor response is consistent with CLIP in the classification experiments on negative queries. Yet OV-DETR also uses CLIP's visual features during training to condition the object proposals~\cite{ovdetr}, but exhibits the best response to negative queries.

\subsection{Is a good classifier all you need for VLM Open-set Recognition?}
\label{sec:goodclassifier}
\begin{figure}[t]
\centering
\begin{subfigure}[b]{0.338\textwidth}
  \centering
  \includegraphics[width=\linewidth]{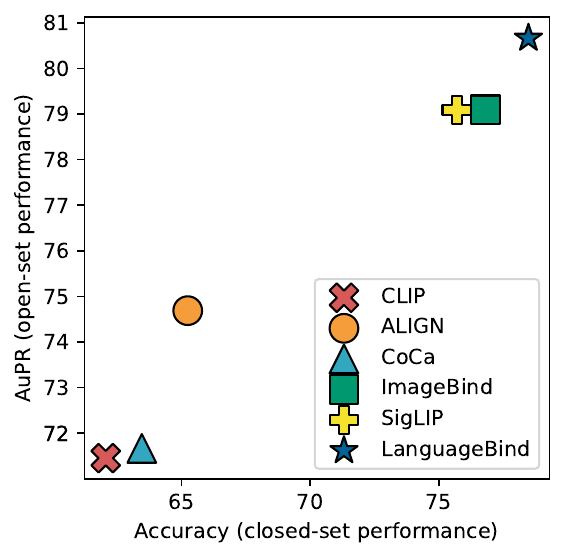}
  \caption{Classification task.}
  \label{fig:sub1}
\end{subfigure}%
\begin{subfigure}[b]{0.35\textwidth}
  \centering
  \includegraphics[width=\linewidth]{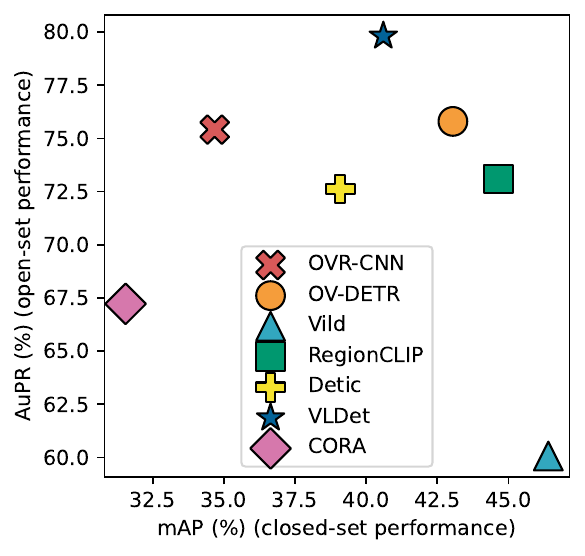}
  \caption{Object Detection task.}
  \label{fig:sub2}
\end{subfigure}
\caption{The relationship between closed-set and open-set performance in VLMs.}
\label{fig:csvsos}
\end{figure}

In traditional open-set recognition, Vaze et. al~\cite{vaze2022open} identified a strong positive correlation between closed-set and open-set performance. Testing this theory for VLM open-set recognition, we compare classifier and object detector closed-set and open-set performance in Fig.~\ref{fig:csvsos}. 

\noindent \textbf{Classification:} 
The correlation observed by~\cite{vaze2022open} appears to hold for VLM classifiers, with greater accuracy (closed-set performance) relating to greater AuPR scores (open-set performance).  
It is worth noting that AuPR is a summary metric and this behaviour may not hold when examining performance at different operating points on the precision-recall curves.

\noindent \textbf{Object Detection:} There is not a clear correlation between a VLM object detector's closed-set and open-set performance. ViLD in particular challenges this correlation, with the greatest mAP yet worst AuPR of all the tested detectors.

\subsection{How Does Query Set Size Impact Performance?}
\label{sec:querysize}
\begin{figure}[t]
    \centering
\includegraphics[width=0.7\linewidth]{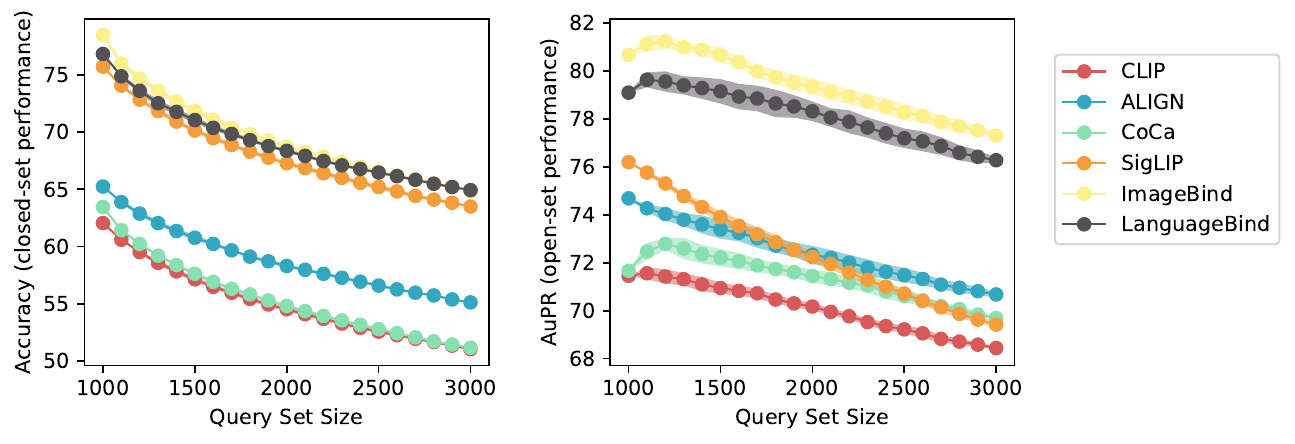}
    \caption{Influence of the query set size on closed-set and open-set performance on ImageNet1k classification, showing mean and standard deviation from 10 random seeds.}
    \label{fig:setsize_cls}
\end{figure}
In Fig.~\ref{fig:setsize_cls}, we test performance of the VLM classifiers on ImageNet while progressively increasing the query set size. We select the additional classes from the WordNet hierarchy by choosing leaf nodes that are related, but not too closely related, to existing ImageNet classes (i.e. share a grandparent in the hierarchy but do not share a direct parent). Increasing the number of target classes and thus the query set size decrements both the closed-set and open-set performance of a VLM classifier. This indicates that testing VLMs with an entire vocabulary of nouns may not be a feasible approach to avoid open-set recognition.

%% file: tables/classification_imagenet_uncertainty.tex
\begin{table}[tb]
\caption{The open-set performance of state-of-the-art VLM classifiers when tested on ImageNet1k. Arrows indicate direction of optimal performance and - indicates operating point could not be achieved.}
\label{tab:cls_unc}
\centering
% \resizebox{\columnwidth}{!}{%
\begin{tabular}{@{}lccccccccc@{}}
\toprule
                            &    & AuROC & AuPR      &  P@95R & R@95P & TP & OSE  & Accuracy \\
Classifier  & Uncertainty & \multicolumn{1}{c}{$\uparrow$} & $\uparrow$ & $\uparrow$ & $\uparrow$ & $\uparrow$     & $\downarrow$         & $\uparrow$              \\  \midrule
\multirow{3}{*}{CLIP \cite{radford2021clip}}  & Softmax & 79.7 &   71.5  &  46.2  & 3.4  &    \multirow{3}{*}{31,023}    & \multirow{3}{*}{50,000} & \multirow{3}{*}{62.1} \\
& Cosine     & 72.2 & 61.3  & 42.5 & 0.1  & & \\
& Entropy    & 80.2 & 72.3 &  45.4  & 2.8  & & \\ 

                            \midrule
\multirow{3}{*}{ALIGN \cite{jia2021align}}  & Softmax & 81.0 & 74.7  &  48.1  & 8.3  &    \multirow{3}{*}{32,618}    & \multirow{3}{*}{50,000} & \multirow{3}{*}{65.2} \\
& Cosine   &  72.4 & 62.1  & 44.8 & - & & \\
& Entropy    & 80.9 & 75.0 &  46.8  & 8.6  & & \\ 
                            \midrule
\multirow{3}{*}{CoCa \cite{yu2022coca}} & Softmax & 79.0 & 71.7 & 46.0 & 3.4 & \multirow{3}{*}{31,725} & \multirow{3}{*}{50,000} & \multirow{3}{*}{63.4}\\
& Cosine & 73.2 & 63.0 & 43.3 & 0.1 \\
& Entropy & 80.5 & 73.0 & 46.8 & 3.5 \\
                            \midrule
\multirow{3}{*}{ImageBind \cite{girdhar2023imagebind}}  & Softmax & 82.8 & 79.1  & 52.8  & 5.0  &    \multirow{3}{*}{38,405}    & \multirow{3}{*}{50,000} & \multirow{3}{*}{76.8} \\
& Cosine  &  79.2 &  73.9  & 50.7 & 0.2  & & \\
& Entropy & 84.3 & 80.1 &  56.3  & 5.6  & & \\ 
\midrule
\multirow{1}{*}{SigLIP \cite{siglip}} & Sigmoid & 81.0 & 76.2 & 52.1 & 2.2 & 37,851 & 50,000 & 75.7 \\

                            \midrule
\multirow{3}{*}{LanguageBind \cite{languagebind}} & Softmax & 83.9 & 80.7 & 54.7 & 10.5 & \multirow{3}{*}{39,243} & \multirow{3}{*}{50,000} & \multirow{3}{*}{78.5}\\
& Cosine & 81.6 & 77.4 & 52.8 & 0.8\\
& Entropy & 85.4 & 81.7 & 58.4 & 11.1\\
                      
                            \bottomrule
\end{tabular}
% }
\end{table}

%% file: tables/detection_coco.tex
% \usepackage{multirow}

\begin{table}[tb]
\caption{The open-set performance of state-of-the-art open-vocabulary object detectors (with ResNet backbone indicated), tested on COCO. Arrows indicate the direction of optimal performance.}
\label{tab:det_unc}
\centering
\resizebox{\columnwidth}{!}{%
\begin{tabular}{@{}lccccccccc@{}}
\toprule
                            &     & Negative & AuPR      &  P@95R & R@95P & TP & OSE                  & mAP                  \\
Detector  & Uncertainty & Embedding & \multicolumn{1}{c}{$\uparrow$} & $\uparrow$ & $\uparrow$ & $\uparrow$     & $\downarrow$         & $\uparrow$              \\ \midrule
\multirow{3}{*}{OVR-CNN (R50) \cite{ovrcnn}}    & Softmax  & Zero-Emb  & 75.4    & 13.0 & 42.9  & \multirow{3}{*}{13,544} & \multirow{3}{*}{190,146}  & \multirow{3}{*}{34.7} \\
    & Cosine  & Zero-Emb & \multicolumn{1}{c}{77.2}       & 15.3 & 43.6 & & \\
    & Entropy  & Zero-Emb   & \multicolumn{1}{c}{63.2}       & 7.5 & 43.4 & & \\ 
\midrule
\multirow{3}{*}{ViLD (R152)\cite{vild}}        & Softmax  & ``background''   & 60.1 & 7.5 & 15.4 & \multirow{3}{*}{16,011} & \multirow{3}{*}{1,485,600} & \multirow{3}{*}{46.4} \\
 & Cosine & ``background'' & 33.1 &  6.9  & 0.1 &  &                          \\
 & Entropy & ``background'' & 62.2 & 8.4 & 16.3 &                      &                         \\ \midrule
\multirow{1}{*}{OV-DETR (R50) \cite{ovdetr}}     & Sigmoid  & None  &    75.8      & 5.9  & 47.3 & 15,818 & 1,485,600 & 43.0 \\ \midrule

\multirow{3}{*}{RegionCLIP (R50)\cite{regionclip}}  & Softmax & Zero-Emb    &   73.1  &  9.9  & 39.4  &    \multirow{3}{*}{15,741}    & \multirow{3}{*}{493,940} & \multirow{3}{*}{44.6} \\
& Cosine  & Zero-Emb    &  74.0  & 7.1 & 40.9  &                      &                          \\
& Entropy & Zero-Emb    & 35.4 &  3.9  & 17.0  &                      &                         \\ \midrule
\multirow{1}{*}{Detic (R50)\cite{detic}}      & Sigmoid & Zero-Emb  &      72.6    &   8.1      & 42.2     &     14,670                 &    495,200                   & 39.1  \\ \midrule
\multirow{1}{*}{VLDet (R50)\cite{vldet}}       & Sigmoid     & Zero-Emb     &   79.8      &  21.9    & 48.5 &      14,751                &   112,923              & 40.6        \\ \midrule
\multirow{1}{*}{CORA (R50)\cite{cora}}       & Sigmoid     &  Zero-Emb   & 67.2    &    8.1    & 24.7   &  13,763                    &   495,200   & 31.6                   \\ 

\bottomrule

\end{tabular}
}
\end{table}

%% file: tables/classification_imagenet_other.tex
\begin{table}[t]
\caption{The open-set performance of state-of-the-art VLM classifiers on ImageNet1k when used with negative random words (R-W) or random embeddings (R-E). 
}
\label{tab:cls_other}
\centering
\resizebox{0.8\columnwidth}{!}{%
\begin{tabular}{@{}lcccccccc@{}}
\toprule
                            &  Negative & AuPR      &  P@95R & R@95P & TP & OSE  & Accuracy \\
Classifier & Embedding & \multicolumn{1}{c}{$\uparrow$} & $\uparrow$ & $\uparrow$ & $\uparrow$     & $\downarrow$         & $\uparrow$              \\ \midrule
& None &   71.5  &  46.2  & 3.4  &  31,023    & 50,000 & 62.1 \\
 % & \cellcolor{gray!25}Simple-W & \cellcolor{gray!25} 71.5 & \cellcolor{gray!25}46.2 & \cellcolor{gray!25}3.4 & \cellcolor{gray!25}31,023 & \cellcolor{gray!25}49,982 & \cellcolor{gray!25}62.0 \\ 
 CLIP \cite{radford2021clip}& \cellcolor{gray!25}500 R-W & \cellcolor{gray!25} 73.4 & \cellcolor{gray!25}48.1 & \cellcolor{gray!25}3.3 & \cellcolor{gray!25}29,961 & \cellcolor{gray!25}41,428 & \cellcolor{gray!25}59.9 \\ 
(Softmax) & \cellcolor{gray!25}2500 R-W & \cellcolor{gray!25} 72.8 & \cellcolor{gray!25}49.8 & \cellcolor{gray!25}1.6 & \cellcolor{gray!25}28,657 & \cellcolor{gray!25}35,092 & \cellcolor{gray!25}57.3 \\ 
 % & Zero-Emb &  71.5 & 46.2 & 3.4 & 31,019 & 49,955 & 62.0 \\ 
& 500 R-E &  71.9 & 47.2 & 2.5 & 30,160 & 44,673 & 60.3 \\ 
& 2500 R-E &  71.7 & 47.1 & 0.9 & 29,291 & 40,497 & 58.6 \\  \midrule
 & None & 74.7  &  48.1  & 8.3  &    32,618  & 50,000 & 65.2 \\
 % & \cellcolor{gray!25}Simple-W & \cellcolor{gray!25} 74.7 & \cellcolor{gray!25}48.1 & \cellcolor{gray!25}8.2 & \cellcolor{gray!25}32,604 & \cellcolor{gray!25}49,832 & \cellcolor{gray!25}65.2 \\ 
ALIGN \cite{jia2021align} & \cellcolor{gray!25}500 R-W & \cellcolor{gray!25} 75.3 & \cellcolor{gray!25}48.9 & \cellcolor{gray!25}10.4 & \cellcolor{gray!25}31,731 & \cellcolor{gray!25}43,198 & \cellcolor{gray!25}63.5 \\ 
(Softmax) & \cellcolor{gray!25}2500 R-W & \cellcolor{gray!25} 74.6 & \cellcolor{gray!25}49.5 & \cellcolor{gray!25}8.9 & \cellcolor{gray!25}30,588 & \cellcolor{gray!25}37,738 & \cellcolor{gray!25}61.2 \\ 
 % & Zero-Emb &  74.7 & 48.1 & 8.3 & 32,618 & 50,000 & 65.2 \\ 
 & 500 R-E &  74.9 & 48.7 & 8.6 & 32,342 & 47,228 & 64.7 \\ 
 & 2500 R-E &  75.1 & 49.7 & 7.2 & 31,462 & 41,210 & 62.9 \\ \midrule

  & None &   71.7  &  46.0  & 3.4 & 31,725 & 50,000  & \multirow{1}{*}{63.4} \\
CoCa \cite{yu2022coca} & \cellcolor{gray!25}500 R-W &   \cellcolor{gray!25}75.4  &  \cellcolor{gray!25}49.6  & \cellcolor{gray!25}0.1  & \cellcolor{gray!25}30,276 & \cellcolor{gray!25}39,528  & \cellcolor{gray!25}60.6 \\
(Softmax) & \cellcolor{gray!25}2500 R-W &   \cellcolor{gray!25}75.9  &  \cellcolor{gray!25}50.9  & \cellcolor{gray!25}6.0  & \cellcolor{gray!25}29,107 & \cellcolor{gray!25}34,360  & \cellcolor{gray!25}58.2 \\
& 500 R-E &  71.7   & 46.0  & 3.4 & 31,723 & 49,966 & 63.4 \\
& 2500 R-E &  71.7   & 46.1  & 3.4 & 31,719 & 49,866  & 63.4 \\

\midrule
 
 & None & 79.1  & 52.8  & 5.0  &    38,405    & 50,000 & 76.8 \\
 % & \cellcolor{gray!25}Simple-W & \cellcolor{gray!25} 79.3 & \cellcolor{gray!25}53.2 & \cellcolor{gray!25}5.7 & \cellcolor{gray!25}38,343 & \cellcolor{gray!25}49,008 & \cellcolor{gray!25}76.7 \\ 
ImageBind \cite{girdhar2023imagebind} & \cellcolor{gray!25}500 R-W & \cellcolor{gray!25} 83.8 & \cellcolor{gray!25}63.1 & \cellcolor{gray!25}7.7 & \cellcolor{gray!25}35,859 & \cellcolor{gray!25}29,172 & \cellcolor{gray!25}71.7 \\ 
 (Softmax) & \cellcolor{gray!25}2500 R-W & \cellcolor{gray!25} 84.6 & \cellcolor{gray!25}66.1 & \cellcolor{gray!25}8.3 & \cellcolor{gray!25}34,179 & \cellcolor{gray!25}22,872 & \cellcolor{gray!25}68.4 \\
 % & Zero-Emb &  79.1 & 52.8 & 5.2 & 38,405 & 49,998 & 76.8 \\ 
 & 500 R-E &  81.0 & 57.3 & 9.0 & 36,966 & 38,636 & 73.9 \\ 
& 2500 R-E &  81.7 & 59.0 & 11.2 & 35,954 & 33,349 & 71.9 \\ 
\midrule

& None &   76.2  & 52.1   & 5.2 & 37,851 & 50,000 &  \multirow{1}{*}{75.7} \\
SigLIP \cite{siglip}& \cellcolor{gray!25}500 R-W &   \cellcolor{gray!25}77.6  &  \cellcolor{gray!25}56.7  & \cellcolor{gray!25}2.3 & \cellcolor{gray!25}36,111  & \cellcolor{gray!25}33,287  & \cellcolor{gray!25}72.2 \\
(Sigmoid) & \cellcolor{gray!25}2500 R-W &   \cellcolor{gray!25}78.8  &  \cellcolor{gray!25}60.2  & \cellcolor{gray!25}2.4 & \cellcolor{gray!25}34,507 & \cellcolor{gray!25}26,110  & \cellcolor{gray!25}69.0 \\
& 500 R-E & 84.7 & 73.0 & 3.6 & 22,772 & 9,570  & 45.5 \\
& 2500 R-E & 90.1 & 83.9 & 7.9 &  10,561 & 2,258  & 21.1  \\
\midrule

  & None &  80.7 & 54.7   & 10.5  & 39,243 & 50,000 & \multirow{1}{*}{78.5} \\
LanguageBind \cite{languagebind} & \cellcolor{gray!25}500 R-W &   \cellcolor{gray!25}84.7  &  \cellcolor{gray!25}64.9  & \cellcolor{gray!25}13.1 & \cellcolor{gray!25}36,596 & \cellcolor{gray!25}28,337  & \cellcolor{gray!25}73.2 \\
(Softmax) & \cellcolor{gray!25}2500 R-W &   \cellcolor{gray!25}85.4  &  \cellcolor{gray!25}68.2  & \cellcolor{gray!25}11.7 & \cellcolor{gray!25}34,540 & \cellcolor{gray!25}21,230  & \cellcolor{gray!25}69.1 \\
& 500 R-E &  80.8   & 54.9  & 10.8 & 39,226 & 49,646  &  78.5 \\
& 2500 R-E &  80.9   &  55.1 & 10.9 & 39,210 & 49,032  & 78.4 \\
\bottomrule
\end{tabular}
}
\end{table}

%% file: tables/detection_coco_other.tex
\begin{table}[t!]
\caption{The open-set performance of state-of-the-art open vocabulary object detectors (with ResNet backbone indicated) when additional negative embeddings are included, tested on COCO. Arrows indicate the direction of optimal performance. Negative embeddings introduced as words are higlighted gray.\cellcolor{gray!25}}
\label{tab:det_other}
\centering
\resizebox{0.95\columnwidth}{!}{%
\begin{tabular}{@{}lcccccccc@{}}
\toprule
                            & Negative  & AuPR      &  P@95R & R@95P & TP & OSE                  & mAP                  \\
Detector & Embedding & \multicolumn{1}{c}{$\uparrow$} & $\uparrow$ & $\uparrow$ & $\uparrow$     & $\downarrow$         & $\uparrow$              \\ \midrule
\multirow{3}{*}{OVR-CNN (R50)\cite{ovrcnn}} & Zero-Emb & 75.4 & 13.0 &  42.9 & 13,544 & 190,146 & 34.7 \\
 & +5 Rand-Embs &  75.6 &  13.1 &  43.1 &  13,543 &  186,120 &  34.7 \\
 & +100 Rand-Embs &  78.3 &  18.2 &  45.6 &  13,129 &  109,267 &  34.0 \\ \midrule
 \multirow{5}{*}{ViLD (R152)\cite{vild}} &\cellcolor{gray!25} ``background'' &\cellcolor{gray!25}  60.1 &\cellcolor{gray!25}  7.5 &\cellcolor{gray!25}  15.4 &\cellcolor{gray!25}  16,011 &\cellcolor{gray!25}  1,485,600 &\cellcolor{gray!25}  46.4 \\
 &\cellcolor{gray!25} +5 Rand-Words &\cellcolor{gray!25}  57.5 & \cellcolor{gray!25} 6.0 & \cellcolor{gray!25} 13.4 &  \cellcolor{gray!25}15,926 & \cellcolor{gray!25} 1,400,612 &  \cellcolor{gray!25}46.6 \\
 &\cellcolor{gray!25} +100 Rand-Words &\cellcolor{gray!25}  55.0 &  \cellcolor{gray!25}2.7 &  \cellcolor{gray!25}10.3 & \cellcolor{gray!25} 15,772 & \cellcolor{gray!25} 1,088,725 & \cellcolor{gray!25} 47.3 \\
 & +5 Rand-Embs &  60.1 &  7.5 &  15.2 &  16,011 &  1,485,231 &  46.4 \\
 & +100 Rand-Embs &  60.2 &  7.6 &  15.2 &  16,019 &  1,478,806 &  46.4 \\ \midrule
\multirow{5}{*}{OV-DETR (R50)\cite{ovdetr}} & None &  75.8 &  5.9 &  47.3 &  15,818 &  1,485,600 &  43.0 \\
 &\cellcolor{gray!25} +5 Rand-Words & \cellcolor{gray!25} 77.2 &  \cellcolor{gray!25}8.1 &  \cellcolor{gray!25}48.3 &  \cellcolor{gray!25}15,510 & \cellcolor{gray!25} 923,398 & \cellcolor{gray!25} 43.0 \\
 &\cellcolor{gray!25} +100 Rand-Words &\cellcolor{gray!25}  84.7 & \cellcolor{gray!25} 27.0 & \cellcolor{gray!25} 55.7 &  \cellcolor{gray!25}13,433 &  \cellcolor{gray!25}239,118 & \cellcolor{gray!25} 40.1 \\
 & +5 Rand-Embs &  76.6 &  6.6 &  47.8 &  15,652 &  1,178,827 &  43.0 \\
 & +100 Rand-Embs &  81.0 &  18.2 &  50.9 &  14,699 &  356,380 &  42.6 \\ \midrule
\multirow{5}{*}{RegionCLIP (R50)\cite{regionclip}}& Zero-Emb &  73.1 &  9.9 &  39.3 &  15,752 &  493,949 &  44.6 \\
 &\cellcolor{gray!25} +5 Rand-Words & \cellcolor{gray!25} 73.8 & \cellcolor{gray!25} 11.1 &  \cellcolor{gray!25}39.7 & \cellcolor{gray!25} 15,662 & \cellcolor{gray!25} 322,596 &\cellcolor{gray!25}  44.6 \\
 & \cellcolor{gray!25}+100 Rand-Words & \cellcolor{gray!25} 77.0 &  \cellcolor{gray!25}17.5 &  \cellcolor{gray!25}43.1 & \cellcolor{gray!25} 15,178 & \cellcolor{gray!25} 154,262 & \cellcolor{gray!25} 44.0 \\
 & +5 Rand-Embs &  91.8 &  77.2 &  44.6 &  8,828 &  5,062 &  27.0 \\
 & +100 Rand-Embs &  99.9 &  99.3 &  83.6 &  287 &  2 &  1.6 \\ \midrule
\multirow{5}{*}{Detic (R50)\cite{detic}} & Zero-Emb &  72.6 &  8.1 &  42.2 &  14,670 &  495,200 &  39.1 \\
 & \cellcolor{gray!25}+5 Rand-Words & \cellcolor{gray!25} 72.8 & \cellcolor{gray!25} 8.5 & \cellcolor{gray!25} 42.4 & \cellcolor{gray!25} 14,637 & \cellcolor{gray!25} 451,633 & \cellcolor{gray!25} 39.1 \\
 & \cellcolor{gray!25}+100 Rand-Words & \cellcolor{gray!25} 74.5 &  \cellcolor{gray!25}11.5 & \cellcolor{gray!25} 43.5 &  \cellcolor{gray!25}14,287 &  \cellcolor{gray!25}265,373 & \cellcolor{gray!25} 38.8 \\
 & +5 Rand-Embs &  93.4 &  74.5 &  61.8 &  10,070 &  6,346 &  30.6 \\
 & +100 Rand-Embs &  100.0 &  100.0 &  100.0 &  371 &  0 &  0.7 \\ \midrule
\multirow{5}{*}{VLDet (R50)\cite{vldet}} & Zero-Emb &  79.8 &  21.9 &  48.4 &  14,751 &  112,954 &  40.6 \\
 & \cellcolor{gray!25}+5 Rand-Words & \cellcolor{gray!25} 80.1 & \cellcolor{gray!25} 22.3 & \cellcolor{gray!25} 48.7 & \cellcolor{gray!25} 14,672 & \cellcolor{gray!25} 111,893 &\cellcolor{gray!25}  40.5 \\
 & \cellcolor{gray!25}+100 Rand-Words & \cellcolor{gray!25} 86.5 & \cellcolor{gray!25} 36.7 & \cellcolor{gray!25} 58.4 & \cellcolor{gray!25} 11,132 & \cellcolor{gray!25} 50,122 & \cellcolor{gray!25} 33.2 \\
 & +5 Rand-Embs &  97.3 &  91.3 &  84.9 &  8,464 &  1,382 &  25.8 \\
 & +100 Rand-Embs &  99.0 &  99.3 &  3.7 &  2,531 &  22 &  5.9 \\ \midrule
\multirow{5}{*}{CORA (R50)\cite{cora}} & Zero-Emb &  67.2 &  8.1 &  24.9 &  13,745 &  495,200 &  31.5 \\
 &\cellcolor{gray!25} +5 Rand-Words & \cellcolor{gray!25} 66.3 &\cellcolor{gray!25}  7.6 &\cellcolor{gray!25}  27.0 & \cellcolor{gray!25} 13,700 & \cellcolor{gray!25} 413,636 & \cellcolor{gray!25} 31.6 \\
 & \cellcolor{gray!25}+100 Rand-Words & \cellcolor{gray!25} 69.1 &\cellcolor{gray!25}  10.1 &\cellcolor{gray!25}  28.6 & \cellcolor{gray!25} 13,722 &  \cellcolor{gray!25}329,393 & \cellcolor{gray!25} 30.1 \\
 & +5 Rand-Embs &  63.1 &  6.2 &  21.8 &  13,689 &  495,200 &  31.5 \\
 & +100 Rand-Embs &  59.9 &  6.0 &  16.7 &  13,680 &  495,174 &  30.6 \\ \bottomrule
\end{tabular}
}
\end{table}

%% file: sections/discussion.tex
\section{Discussion}
\label{sec:discussion}

\noindent\textbf{Open Challenges:}
The tested baseline uncertainty measures were ineffective at identifying open-set errors, indicating a need to research better alternatives. While uncertainty representations for VLMs have been proposed~\cite{ji2023map, upadhyay2023probvlm}, their utility for open-set recognition remains unexplored. 

While the use of negative random embeddings was able to reduce open-set error, large numbers were typically required to have any effect and this was often at the cost of falsely captured true positive predictions.
Interestingly, some negative embeddings were much more effective at capturing open-set errors than others (e.g. the random string ``awy'' tested with ImageBind, see the Supplementary Sec. S9). Identifying the characteristics of these embeddings, learning them through prompt-tuning~\cite{promptdet, detpro, cho2023distribution, zhu2023prompt, li2024learning} or introducing auto-labelling techniques~\cite{zara2023autolabel} are promising directions.
 
\noindent\textbf{Conclusions:}
VLMs are increasingly applied in diverse real-world applications, from robotics to medical image analysis. We hope that by highlighting their weaknesses, we encourage further research to improve these models and enhance their positive societal impact. We have clearly demonstrated that VLMs for open-vocabulary image classification and object detection introduce closed-set assumptions by relying on a finite query set. We thus refuted the assumption that vision-language models (VLMs) are inherently open-set models due to their training on extensive internet-scale datasets. While the investigated baseline approaches to identify open-set errors showed promise, we found none of them sufficiently robust and reliable for safety-critical applications. Our new benchmark and evaluation protocol is designed to foster and support this research. Understanding VLMs' limitations ultimately reduces risks, improves transparency, and ensures safer deployment in critical applications.

%% file: sections/supplementary.tex
\newpage 
\makeatletter
\renewcommand \thesection{S\@arabic\c@section}
\renewcommand\thetable{S\@arabic\c@table}
\renewcommand \thefigure{S\@arabic\c@figure}
\makeatother

\title{Supplementary Material}
\author{}
\institute{}
% \author{Dimity Miller \and
% Niko S\"underhauf \and
% Alex Kenna \and
% Keita Mason
% }

% \institute
\maketitle

\setcounter{section}{0}
\setcounter{figure}{0}
\setcounter{table}{0}

\section{Additional Evaluation Details}
We have made our testing and evaluation repository publicly available at \href{http://github.com/dimitymiller/openset\_vlms}{\url{github.com/dimitymiller/openset\_vlms}}.
We use the \href{https://github.com/cocodataset/cocoapi/tree/master}{COCO API} to identify True Positive (TP) predictions in object detection (see Sec.~\ref{sec:protocol_data}) -- this is the established process for identifying TPs used during mAP calculation~\cite{lin2014microsoft_supp}. 
% It can be seen in \texttt{detection/scripts/evaluate.py} \texttt{L128-180} of our attached code repository.

% -- a TP correctly predicts the object class and overlaps with an Intersection over Union (IoU) greater than $0.5$, with object assignment preference based on detection confidence. All predictions that do not meet these conditions are false positives (FPs). We collect the confidence (or uncertainty) associated with all TPs for later evaluation with the performance metrics.

\section{Model Implementation Details}
We list the testing configuration used for the VLM classifiers in the Table below. This can also be seen in the evaluation repository included in the supplementary folder, where we include the python scripts used to test and collect results from each of the VLM classifiers.

\input{tables/vlm_cls_impl}

We list the testing configuration used for each VLM detectors in the Table below. We use the identified config files or testing scripts without changes unless specified otherwise. For all detectors, we do not impose any minimum confidence threshold for predictions -- low confidence predictions are naturally handled by our AuPR metric which considers confidence thresholds. 

\input{tables/vlm_det_impl}
\newpage
\section{ROC Curve Metrics}
The area under the ROC curve (AuROC) is a popular metric in the traditional open-set recognition literature~\cite{chen2021adversarial, oza2019c2ae, perera2020generative, vaze2022open, zhang2020hybrid}. However a key limitation of ROC curves is that they are not affected by the size of the negative class~\cite{saito2015precision} -- this can make them less informative and even misleading when used on datasets where there is a large imbalance between the positive and negative class. We highlight this in
% Tab.~\ref{tab:cls_unc_auroc} and
Tab.~\ref{tab:det_auroc} for VLM object detection -- VLM detectors with \emph{larger} numbers of OSE achieve greater AuROC  performance than detectors with \emph{lower} OSE counts. Due to these shortcomings of AuROC, we prefer AuPR for the discussion of experiments in the main paper.

\input{tables/detection_coco_auroc}

% We can construct precision-recall (PR) curves by varying an uncertainty threshold and calculating the precision and recall at each point \cite{}. Predictions with uncertainty below the threshold are classed as the positive class (i.e. kept as likely TPs), otherwise predictions are classed as the negative class (i.e. rejected as OSEs). A desirable quality of PR curves is they are affected by the total numbers of the negative class (OSEs) as well as the uncertainty of all predictions. A high precision indicates that little-to-none OSEs remain after the uncertainty thresholding. A high recall indicates that most TP predictions have been kept and not incorrectly rejected during thresholding. We summarise the performance of the precision-recall curves with 3 metrics: (1) \textbf{Area under the PR curve (AuPR)}, (2) \textbf{Precision at $\mathbf{95\%}$ Recall}, and (3) \textbf{Recall at $\mathbf{95\%}$ Precision}. 

\newpage
\section{Domain-specific Image Classification Results}
We test three VLM classifiers -- CLIP~\cite{radford2021clip}, ALIGN~\cite{jia2021align} and ImageBind~\cite{girdhar2023imagebind} -- on an additional three popular domain-specific classification datasets: 

\begin{enumerate}
    \item German Traffic Sign Recognition Benchmark (GTSRB)~\citeSM{gtsrb}: containing images of 34 different types of traffic signs. We use the same text prompts to describe each traffic sign as CLIP~\cite{radford2021clip}, see \href{https://github.com/openai/CLIP/blob/main/data/prompts.md}{here}. We test on the test split of the dataset, which contains $12,630$ images.
    \item Places365-Standard (Places365)~\citeSM{places365}: containing images of 365 different scene categories. We use the category labels directly as text prompts (e.g. ``amusement park'', ``aquarium'', etc.). We test on the validation split of the dataset, which contains $36,500$ images.
    \item Food101~\citeSM{food101}: containing images of 101 different food categories. We use the category labels directly as text prompts (e.g. ``apple pie'', ``bibimbap'', etc.). We test on the test split of the dataset, which contains $25,250$ images.
\end{enumerate}

\begin{figure}[th!]
\centering
\includegraphics[width=0.7\linewidth]{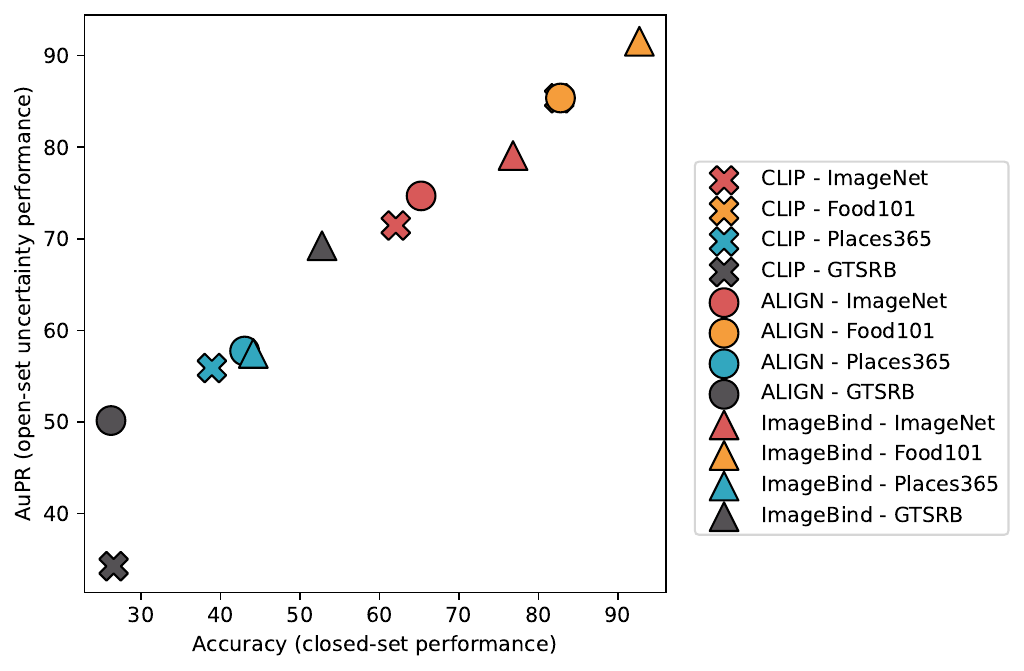}
\caption{Results for the domain-specific datasets when investigating the relationship between closed-set and open-set performance.}
\label{fig:csvsos_domain}
\end{figure}

In Fig.~\ref{fig:csvsos_domain}, we reproduce our experiments on the relationship between closed-set and open-set performance from Sec.~\ref{sec:goodclassifier}. %Sec 5.3
% on the impact of the query set size on the closed-set and open-set performance of the VLM. Consistent with Fig.~\ref{fig:setsize_cls}, increasing query set sizes decrement the closed-set performance. Generally the open-set performance also decrements with increasing query set size, although this is not the case on the GTSRB dataset -- this highlights that there may be domain-specific trends relating to open-set performance that warrant further investigation by the community.

% \begin{figure}[th!]
% \centering
% \begin{subfigure}[t]{0.8\textwidth}
%   \centering
%     \includegraphics[width=\linewidth]{figures/setsize_classification_gtsrb.pdf}
% \end{subfigure}

% \begin{subfigure}[t]{0.8\textwidth}
% \centering
% \includegraphics[width=\linewidth]{figures/setsize_classification_places365.pdf}
% \end{subfigure}

% \begin{subfigure}[t]{0.8\textwidth}
% \centering
% \includegraphics[width=\linewidth]{figures/setsize_classification_food101.pdf}
% \end{subfigure}
% \caption{Results for the domain-specific datasets when investigating the influence of query set size on closed-set and open-set performance.}
% \label{fig:domain_setsize}
% \end{figure}

\newpage
\section{Uncertainty for Open-set and Closed-set Errors}
We compare uncertainty performance for identifying errors in open-set recognition (i.e. open-set errors) versus standard closed-set recognition (i.e. closed-set misclassifications). We report the softmax and sigmoid uncertainty baselines for the VLM classifiers and object detectors. 

Tab.~\ref{tab:cls_unc_csvsos} shows the results for VLM classification, where the performance of uncertainty clearly degrades for open-set error compared to closed-set error for all six classifiers. In particular, thresholding for high precision of predictions (i.e. low error rates) shows substantial performance differences, indicating there may be a greater portion of overconfident open-set errors than closed-set errors. Interestingly, SigLIP with the sigmoid activation and uncertainty appears to show the least discrepancy between closed-set error and open-set error performance; In fact, the AuROC performance for open-set errors is better than for closed-set error. This suggests that sigmoid uncertainty may not be a powerful indicator for closed-set error identification.

\input{tables/classification_csvsos}

When performing the closed-set test with the VLM object detectors, false positives can be categorised as either closed-set error or open-set error. False positives that are caused by ``background'' detections -- detections that do not overlap a labelled object from the dataset class list -- are actually open-set errors under our proposed definition. False positives that are caused by misclassification -- detections that overlap a labelled object from the dataset class list but classify incorrectly -- are closed-set errors. For this experiment, we identify closed-set errors as false positives from the closed-set test that overlap a labelled object with an IoU greater than $0.1$ but misclassify the object class.

Tab.~\ref{tab:det_unc_csvsos} shows the results for VLM detection. Similar to the classification results, the performance of uncertainty clearly degrades for open-set error compared to closed-set error for all seven object detectors. For some detectors, the performance difference is substantial (e.g. the AuPR of CORA decrements by 10.2\% between closed-set and open-set performance), whereas others are less notable (e.g. OV-DETR). For all detectors, one of the primary degradations from closed-set to open-set is the number of errors, with between $1.8\times$ to $11.7\times$ more open-set than closed-set misclassifications depending on the detector.

\input{tables/detection_csvsos}

\newpage
\section{Qualitative Examples}
Below we show a selection of images containing open-set errors from each of the VLM object detectors. We threshold predictions from each detector at its individual $95\%$ precision confidence threshold to show the most confident and pervasive open-set errors. 
% Some open-set objects are semantically close to the predicted class (e.g. cars predicted as trucks), although others are not (e.g. rooster predicted as horse, or stool predicted as wine glass).

\begin{figure}[h!]
\centering
\captionsetup[subfigure]{labelformat=empty}
\begin{subfigure}[b]{\textwidth}
  \centering
  \includegraphics[width=0.4\linewidth]{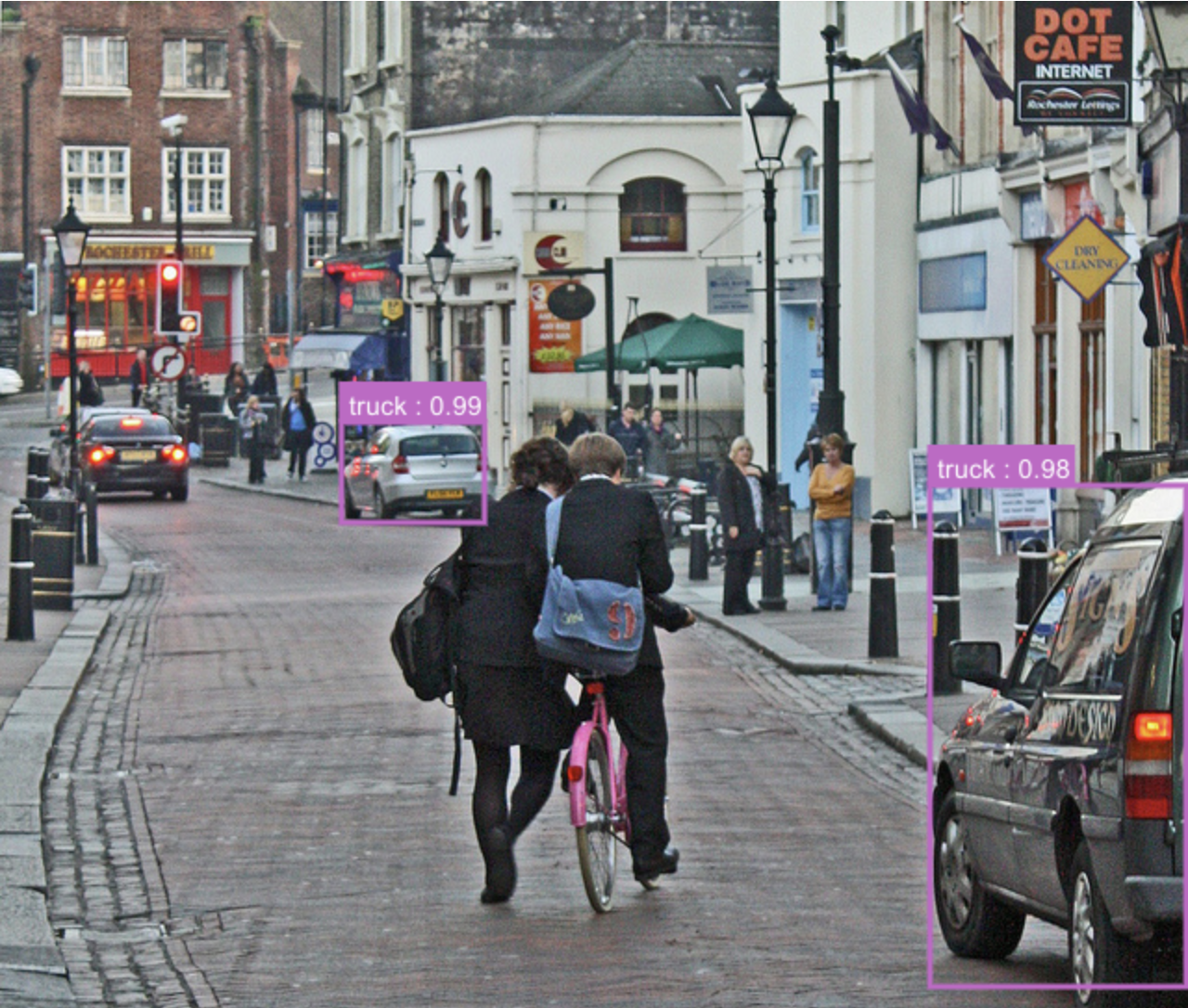}
  \includegraphics[width=0.59\linewidth]{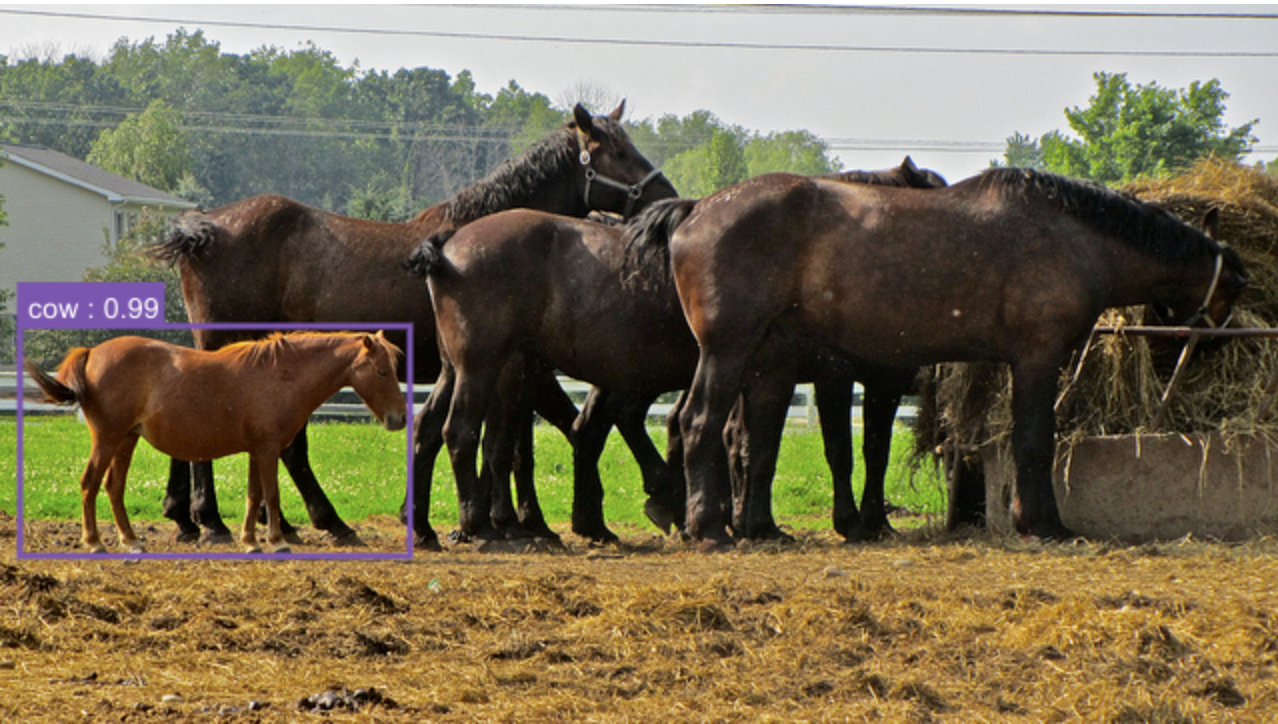}
\subcaption{OVR-CNN~\cite{ovrcnn}.}
\end{subfigure}

\begin{subfigure}[b]{\textwidth}
  \centering
  \includegraphics[width=0.4\linewidth]{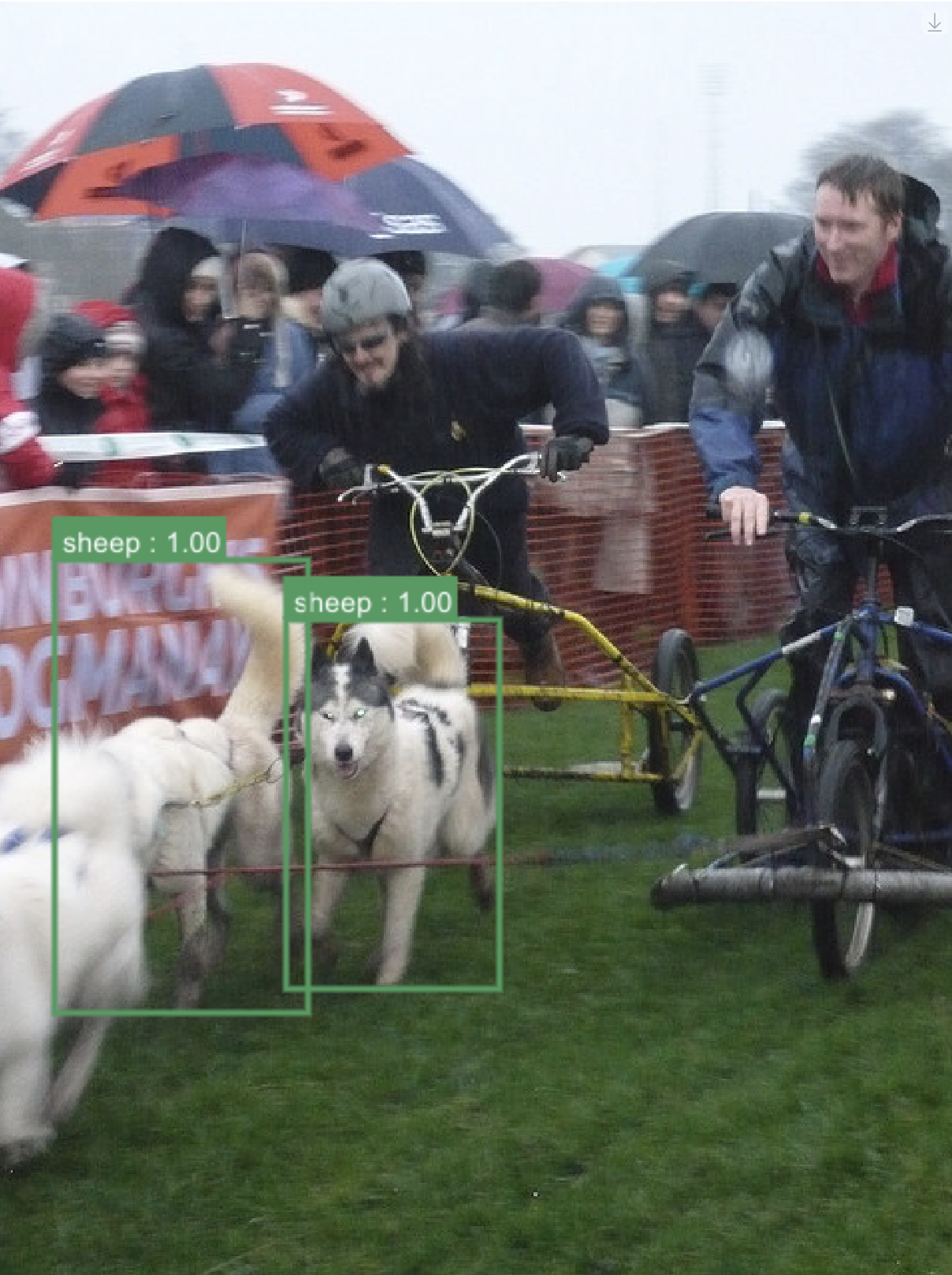}
  \includegraphics[width=0.35\linewidth]{figures/vild_3.png}
\subcaption{ViLD~\cite{vild}.}
\end{subfigure}

\begin{subfigure}[b]{\textwidth}

  \centering
  \includegraphics[width=0.54\linewidth]{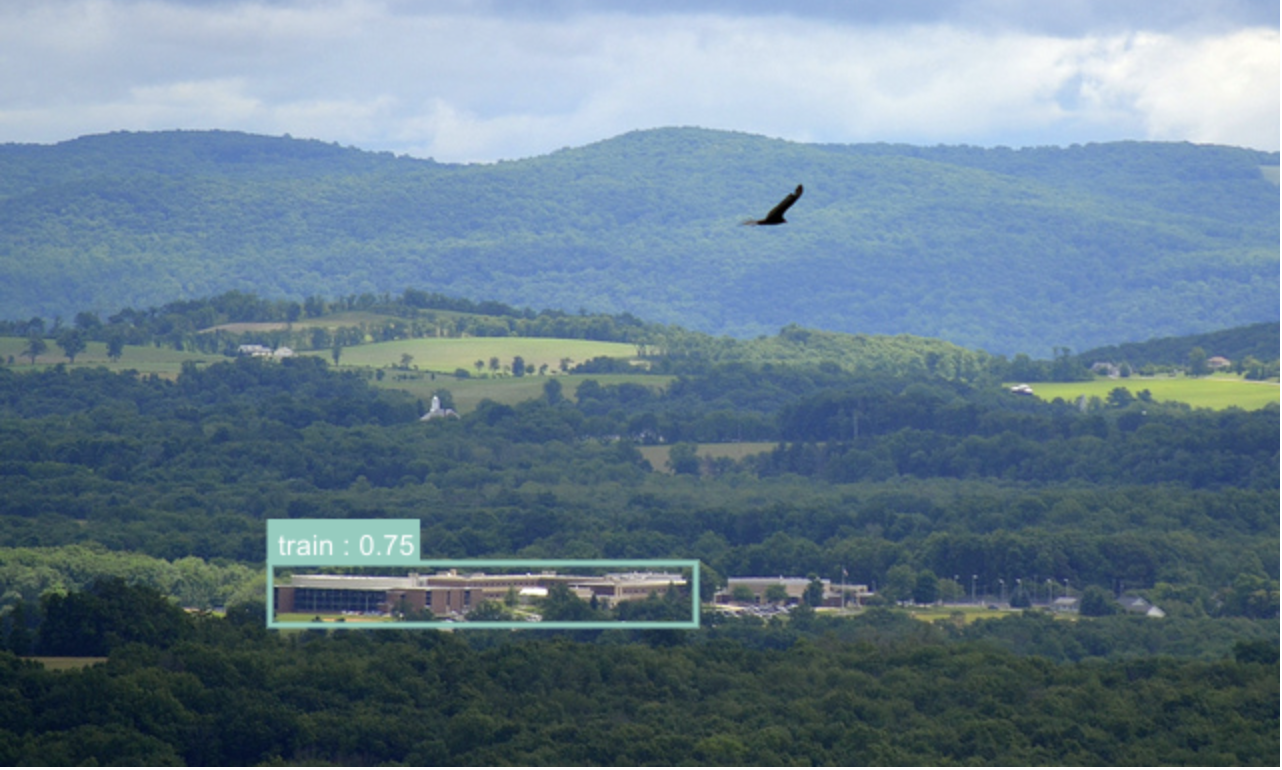}
  \includegraphics[width=0.44\linewidth]{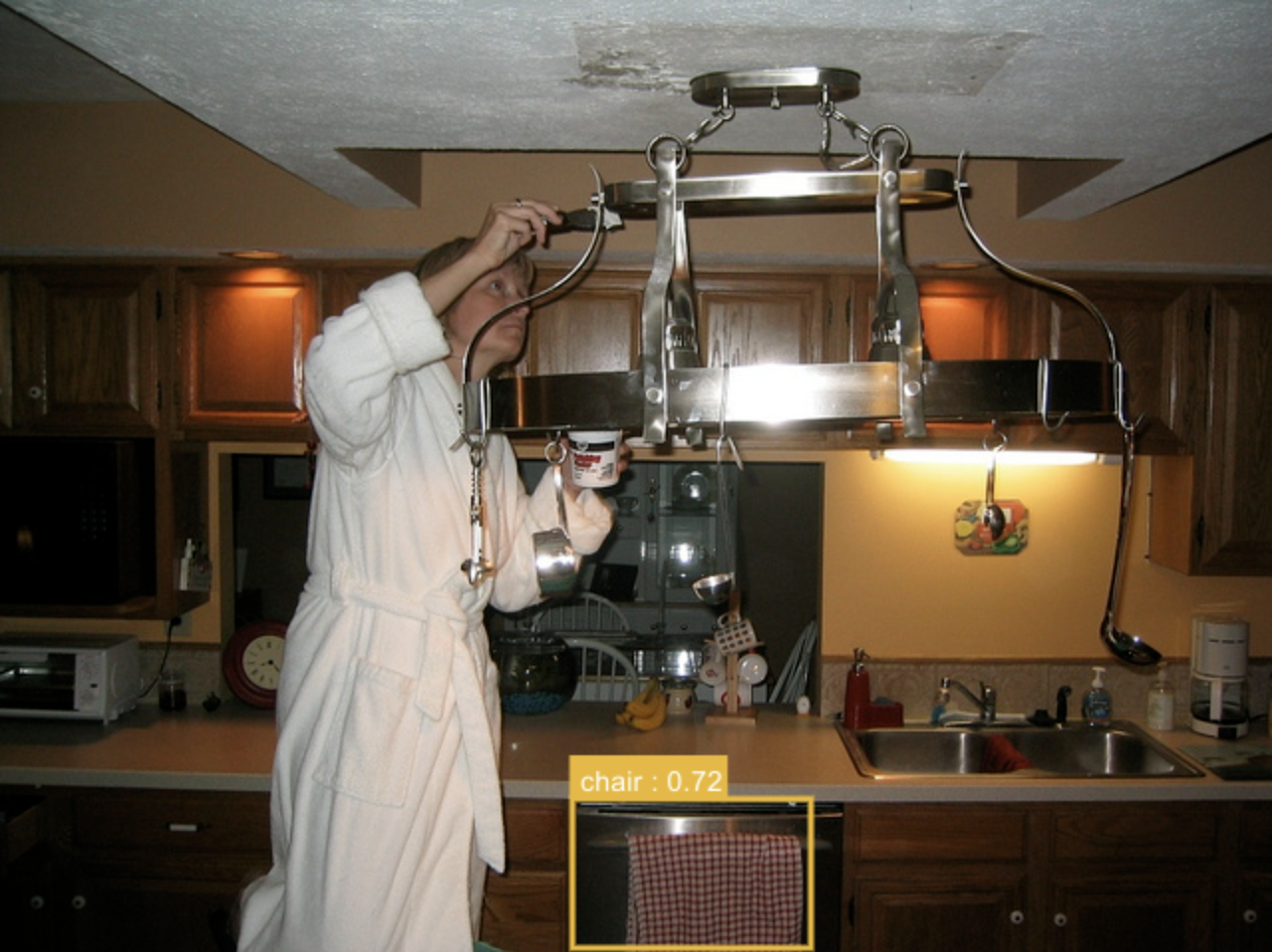}
\subcaption{OV-DETR~\cite{ovdetr}.}
\end{subfigure}
\end{figure}

\begin{figure}[h!]
\centering
\captionsetup[subfigure]{labelformat=empty}
\begin{subfigure}[b]{\textwidth}
  \centering
  \includegraphics[width=0.48\linewidth]{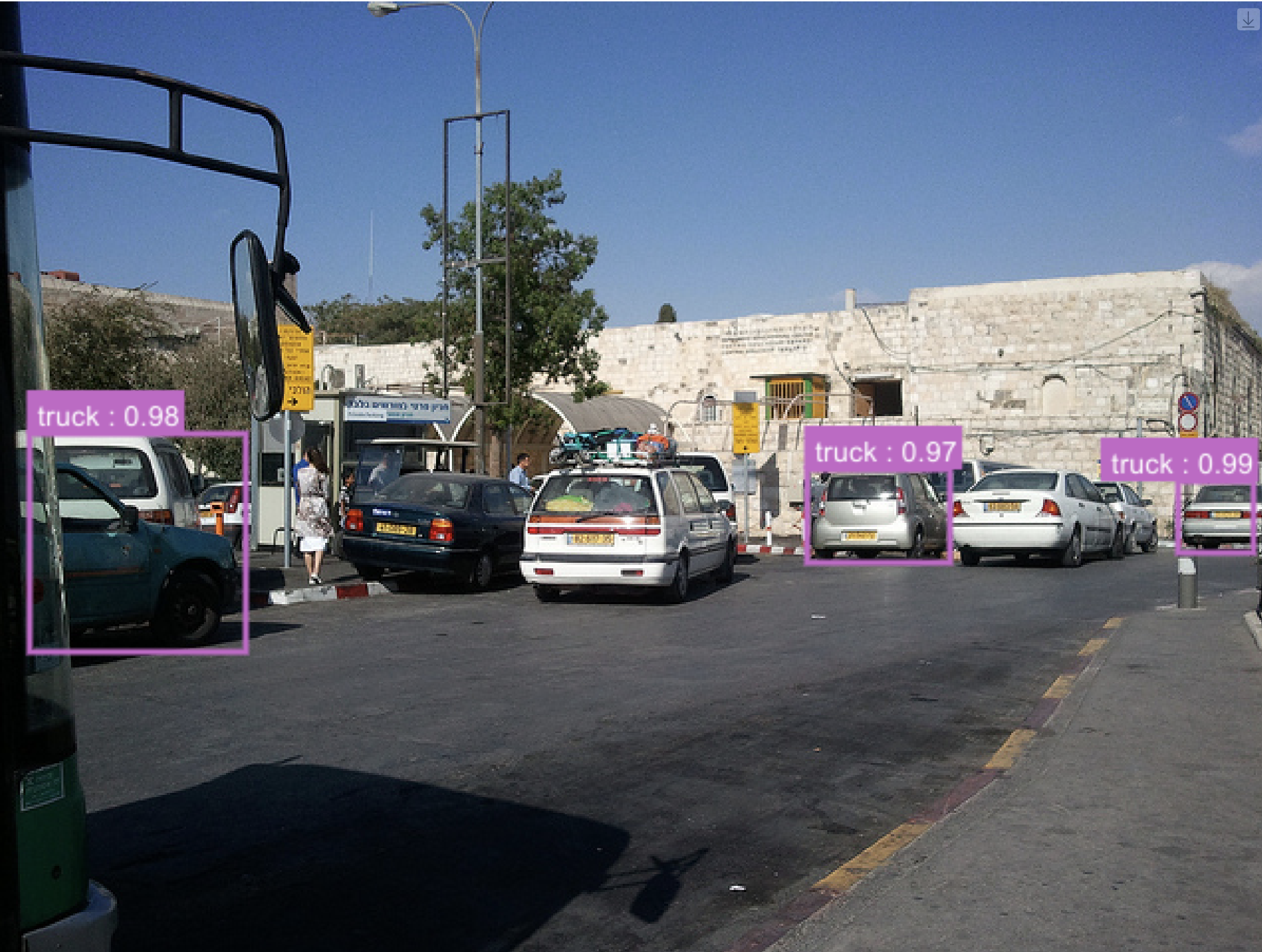}
  \includegraphics[width=0.48\linewidth]{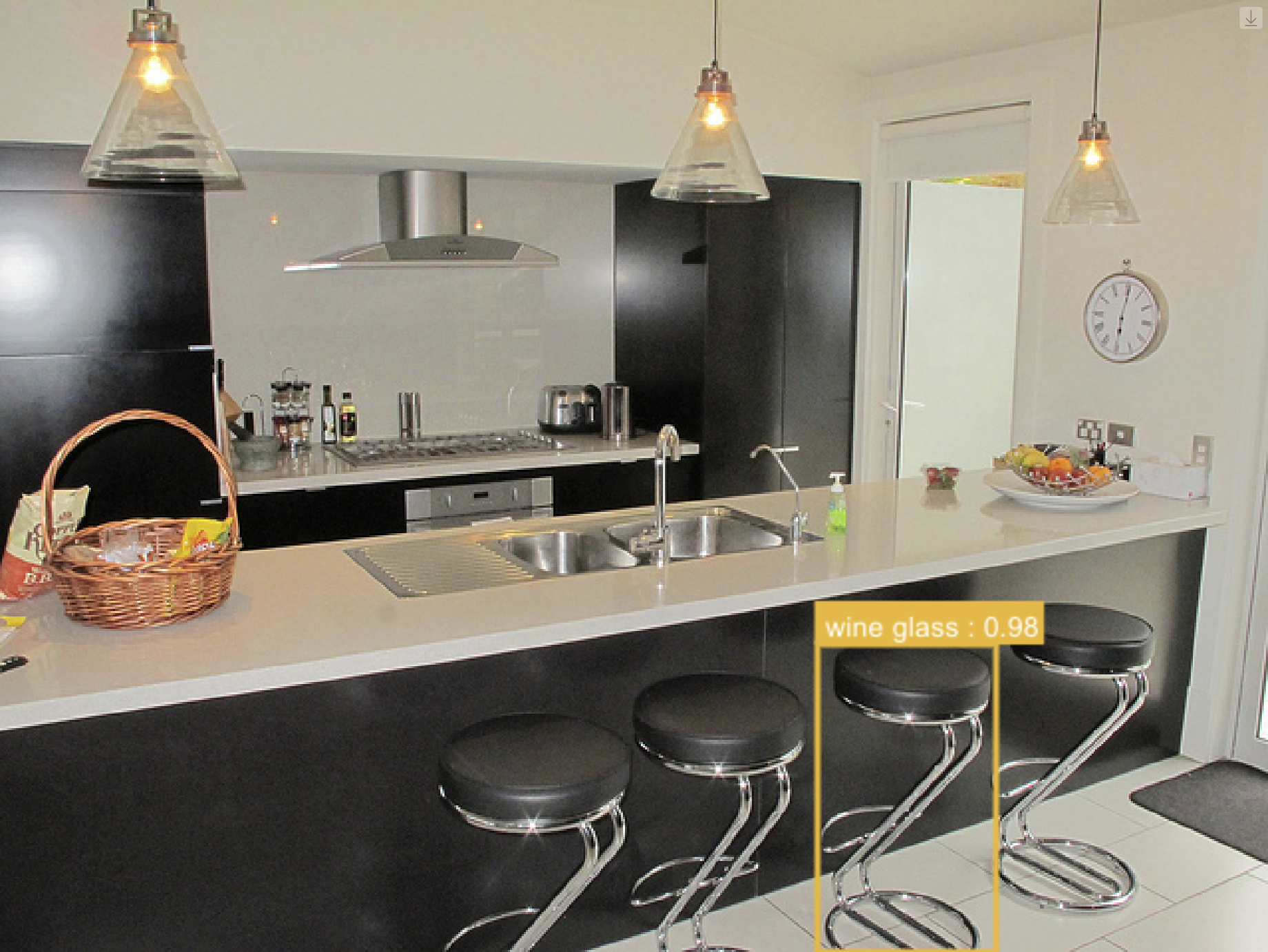}
\subcaption{RegionCLIP~\cite{regionclip}.}
\end{subfigure}

\begin{subfigure}[b]{\textwidth}
  \centering
  \includegraphics[width=0.39\linewidth]{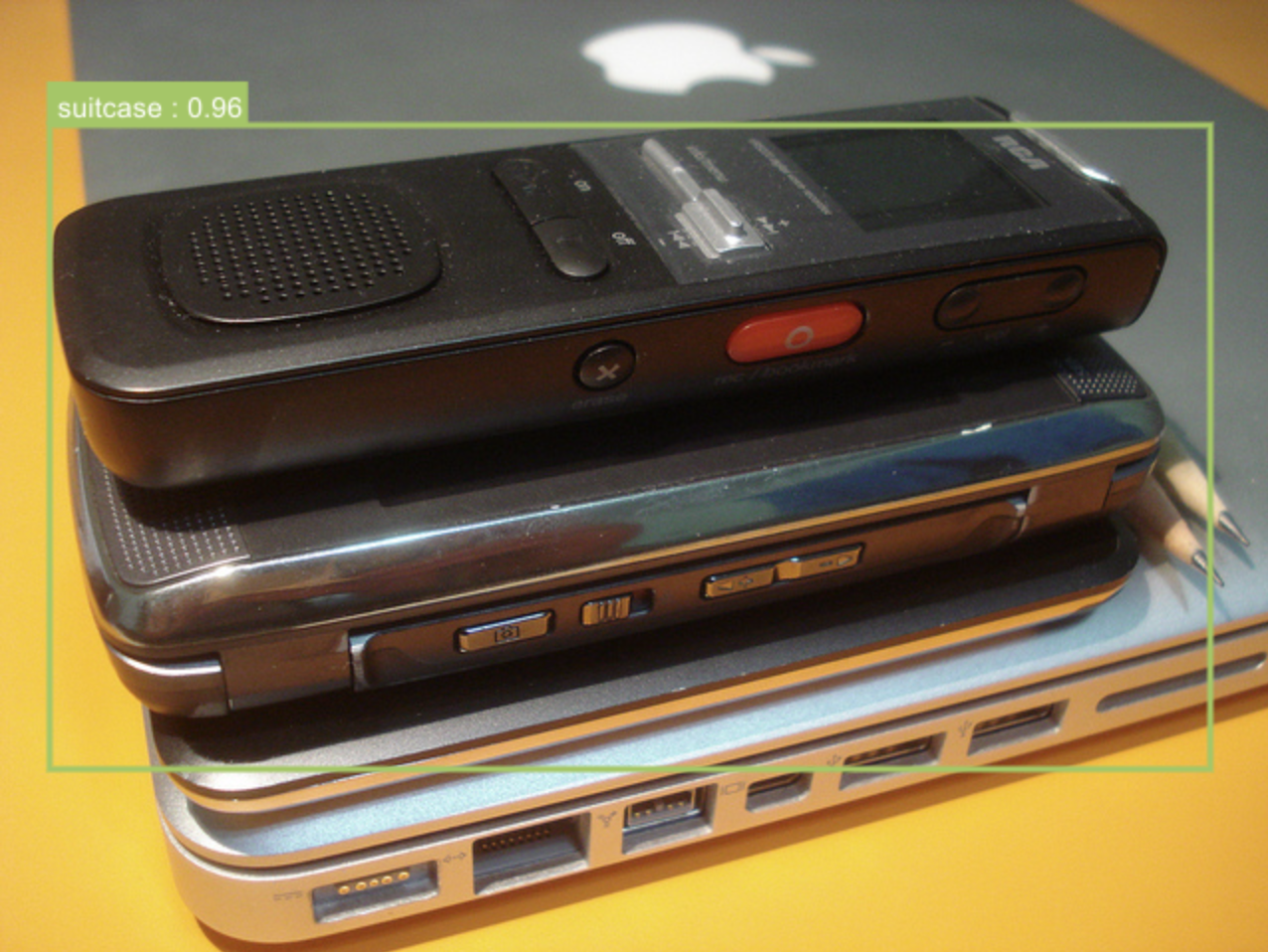}
  \includegraphics[width=0.5\linewidth]{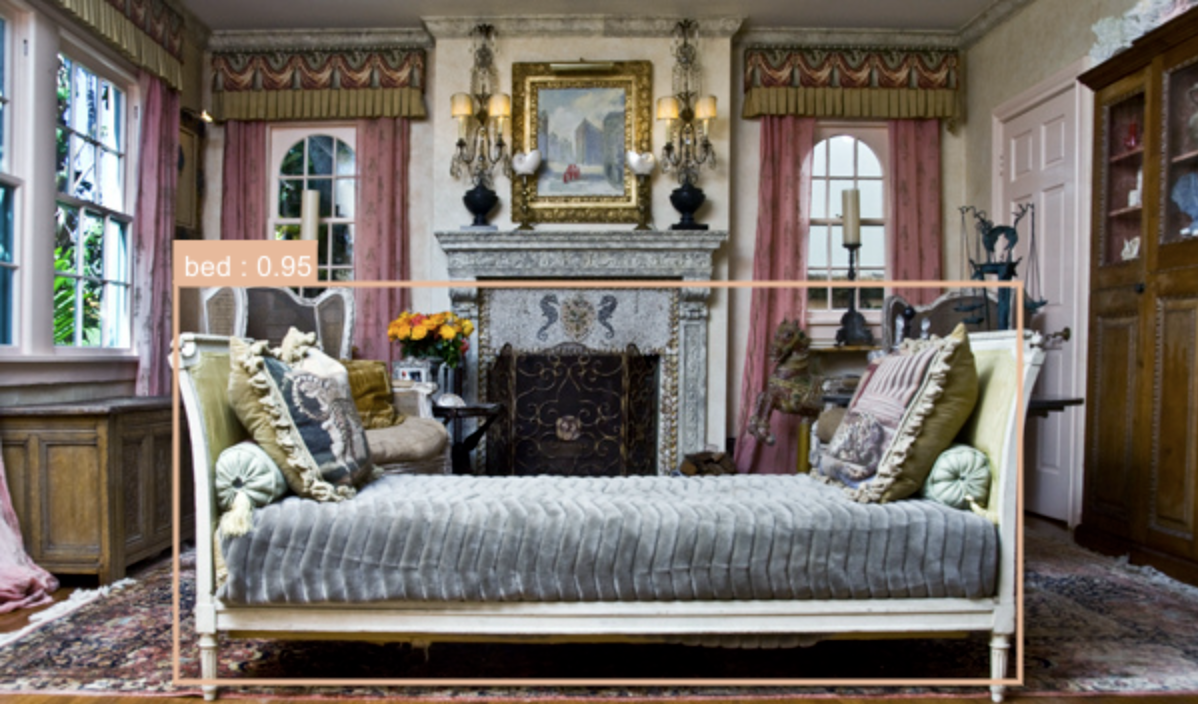}
\subcaption{Detic~\cite{detic}}
\end{subfigure}

\begin{subfigure}[b]{\textwidth}
  \centering
  \includegraphics[width=0.45\linewidth]{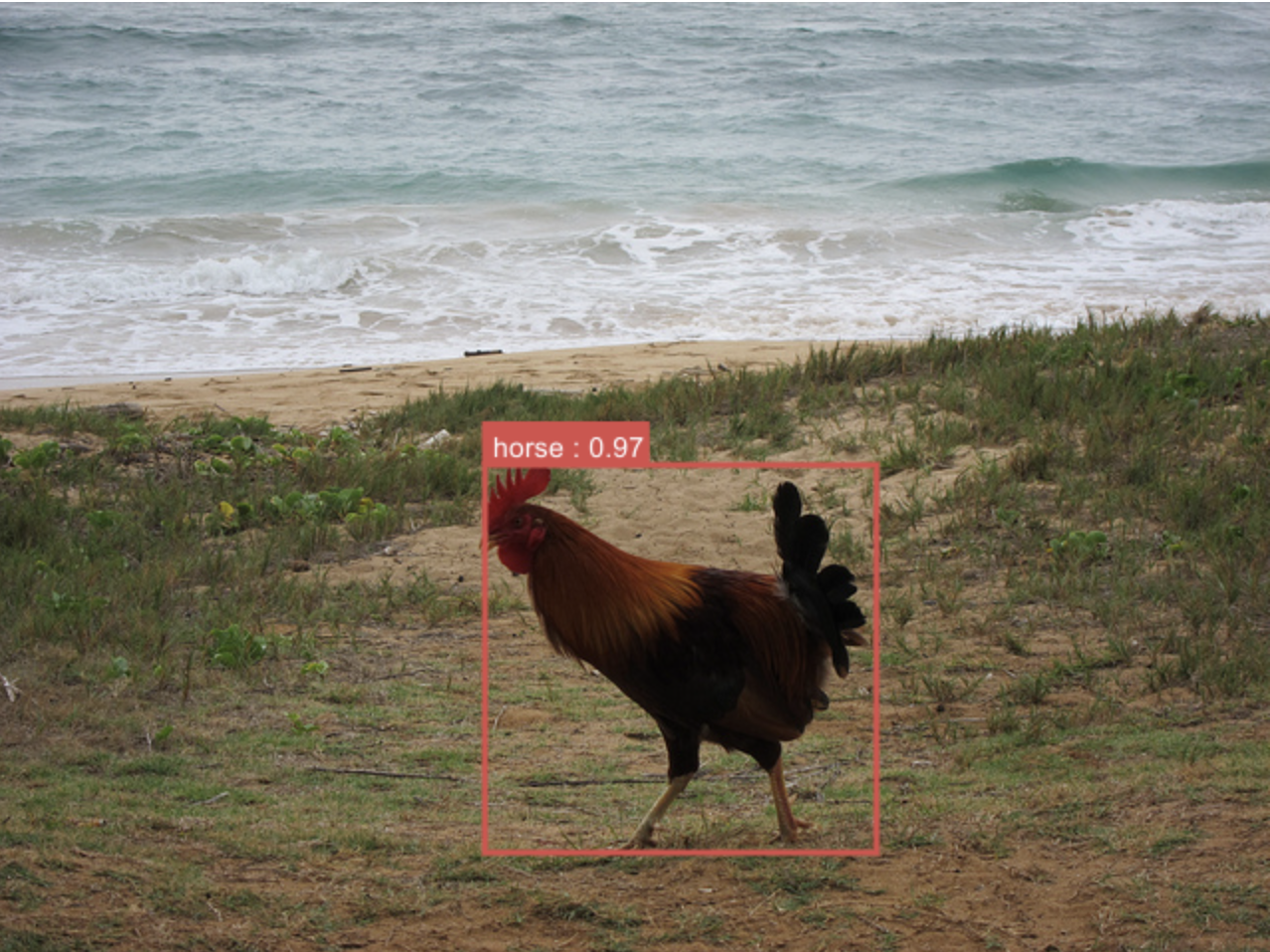}
  \includegraphics[width=0.5\linewidth]{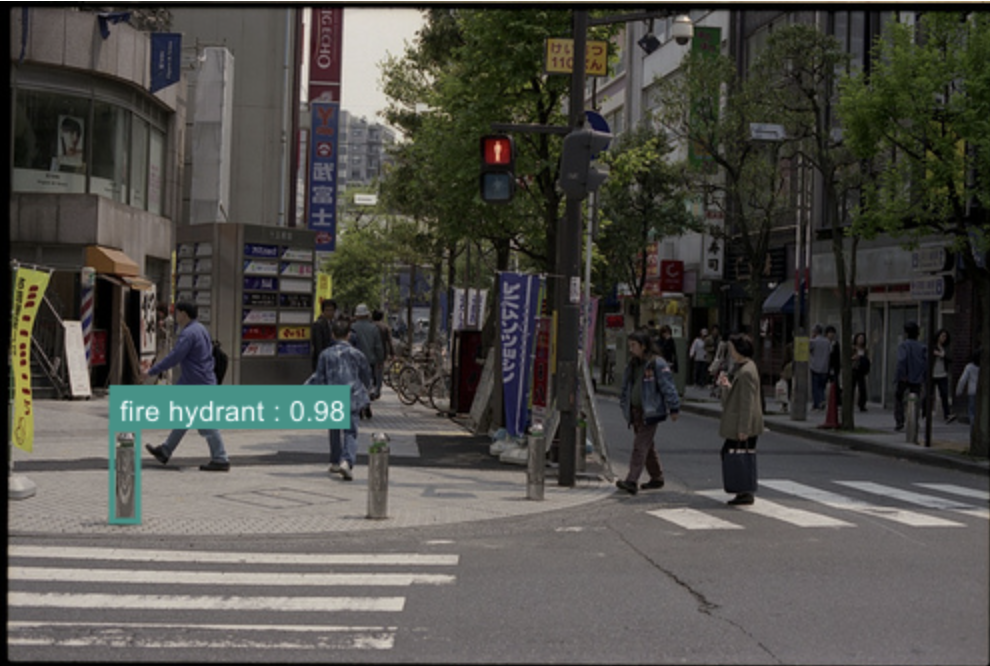}
\subcaption{VLDet~\cite{vldet}.}
\end{subfigure}

\begin{subfigure}[b]{\textwidth}
  \centering
  \includegraphics[width=0.47\linewidth]{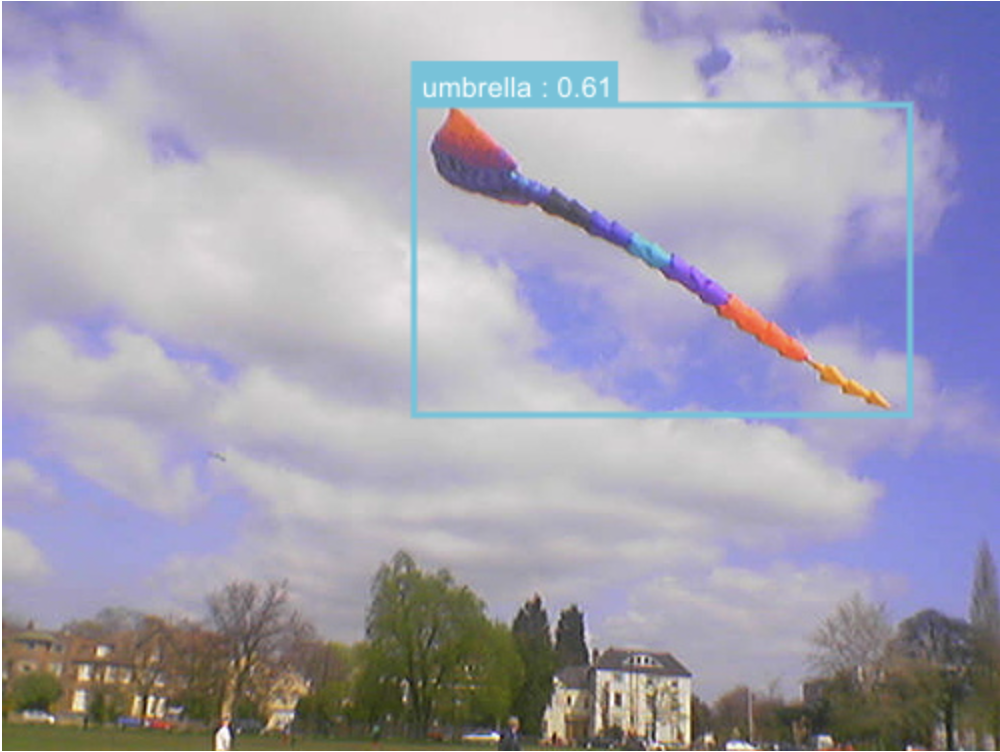}
  \includegraphics[width=0.49\linewidth]{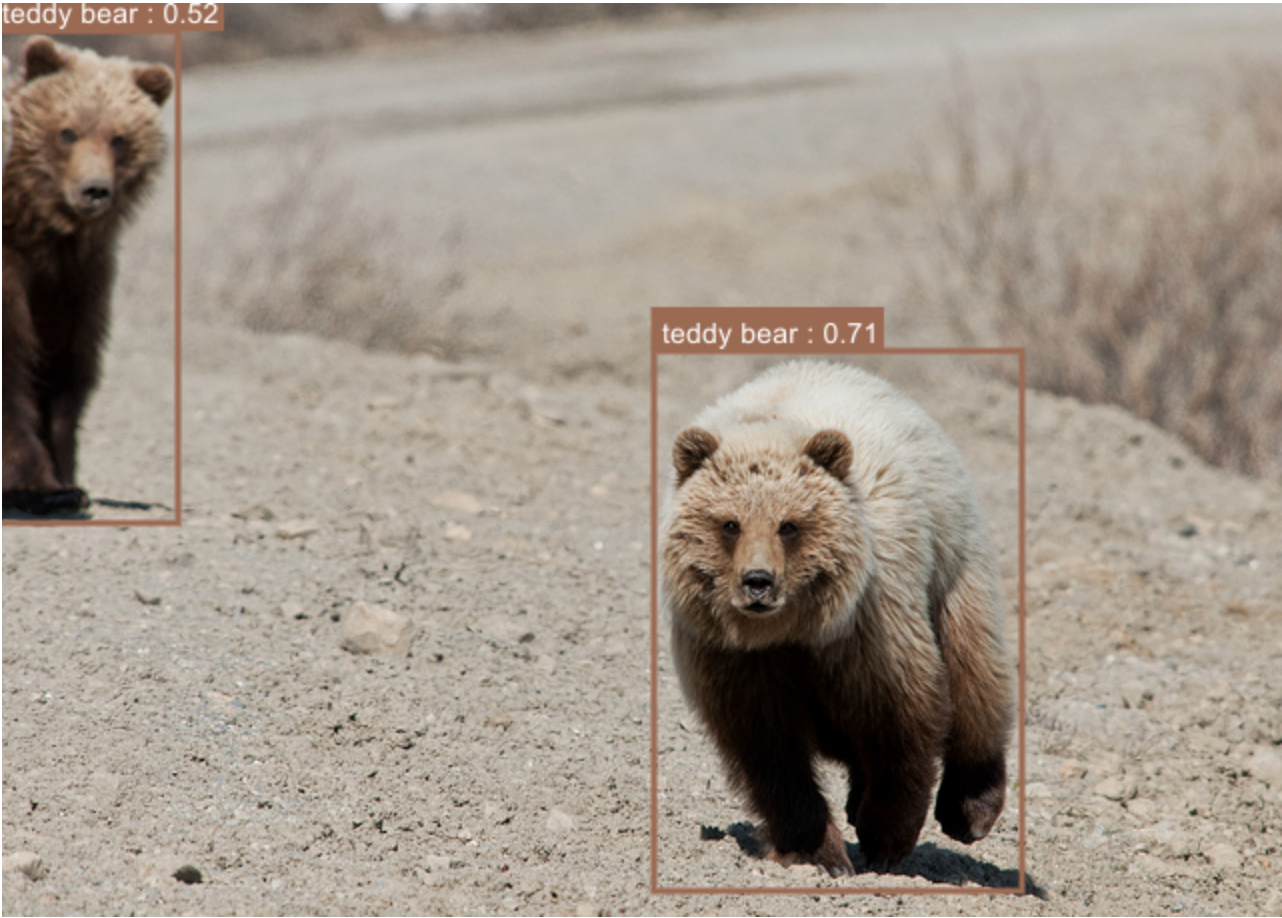}
\subcaption{CORA~\cite{cora}.}
\end{subfigure}
\caption{Qualitative examples of open-set errors from the VLM object detectors.}
\label{qual_examples}
\end{figure}

\clearpage
\section{Uncertainty Histograms}
\begin{figure}[th!]
\centering
\includegraphics[width = 0.8\linewidth]{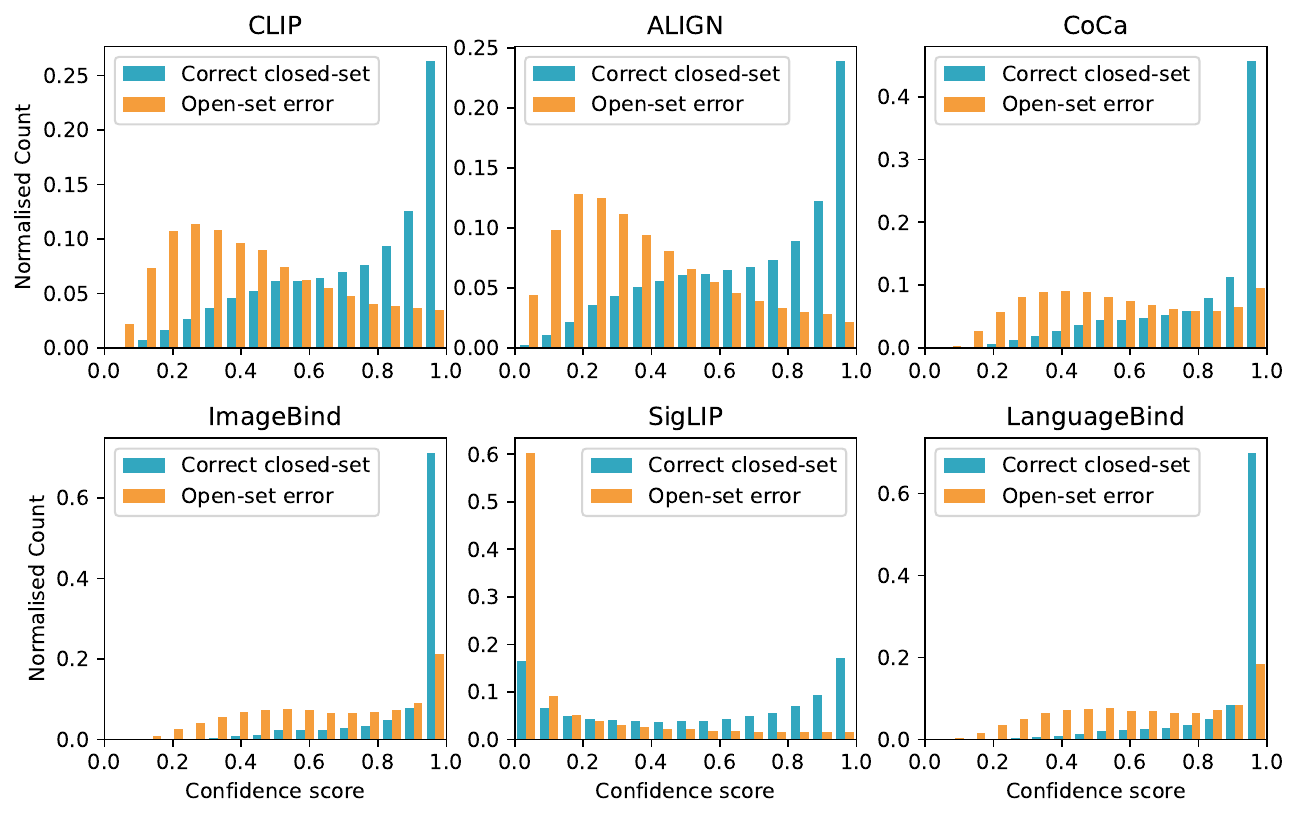}
\caption{The VLM classifiers suffer from overconfident open-set errors, i.e. open-set errors with very high softmax scores.}
\label{cls_unchist}
% \end{figure}

% \begin{figure}[t]
\centering
\begin{subfigure}[b]{0.24\textwidth}
  \centering
  \includegraphics[width=\linewidth]{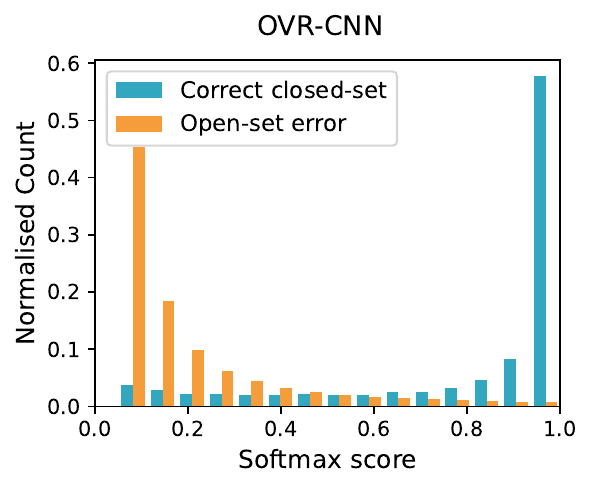}
\end{subfigure}%
\begin{subfigure}[b]{0.24\textwidth}
  \centering
  \includegraphics[width=\linewidth]{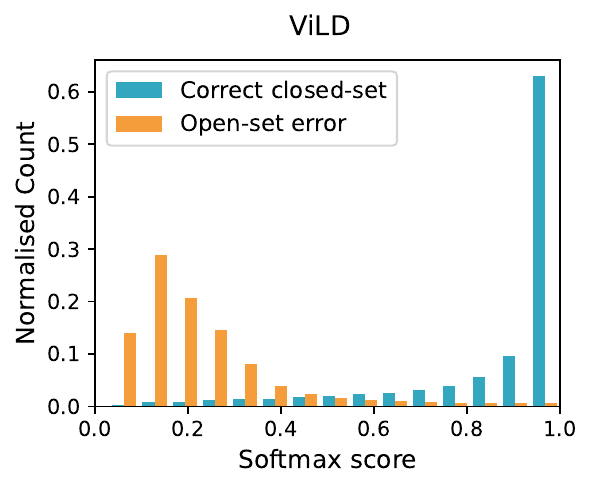}
\end{subfigure}
\begin{subfigure}[b]{0.24\textwidth}
  \centering
  \includegraphics[width=\linewidth]{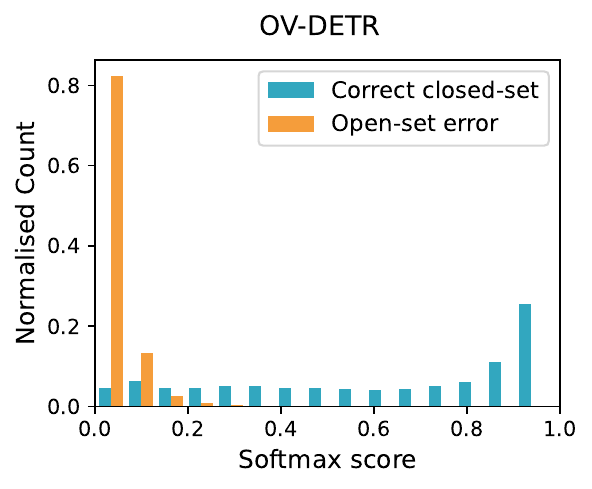}
\end{subfigure}
\begin{subfigure}[b]{0.24\textwidth}
  \centering
  \includegraphics[width=\linewidth]{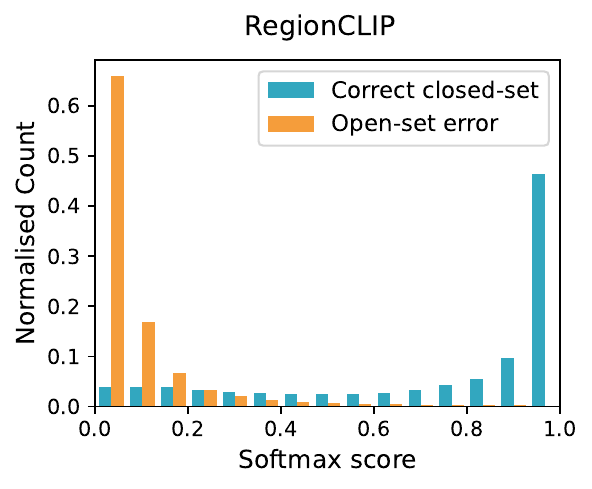}
\end{subfigure}

\begin{subfigure}[b]{0.24\textwidth}
  \centering
  \includegraphics[width=\linewidth]{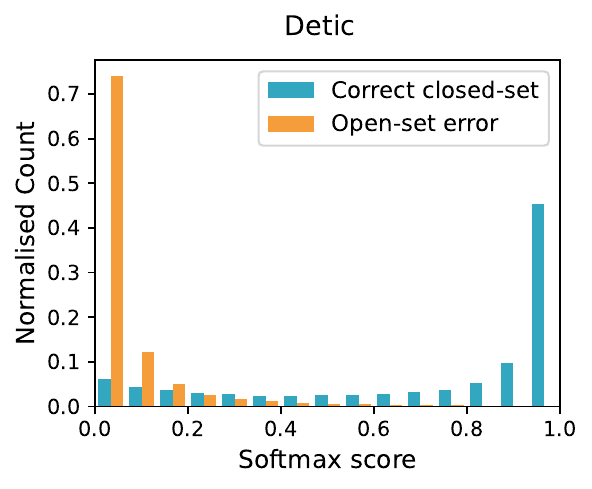}
\end{subfigure}
\begin{subfigure}[b]{0.24\textwidth}
  \centering
  \includegraphics[width=\linewidth]{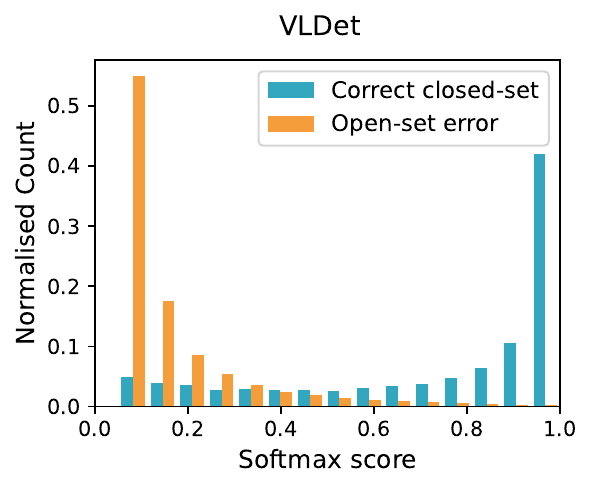}
\end{subfigure}
\begin{subfigure}[b]{0.24\textwidth}
  \centering
  \includegraphics[width=\linewidth]{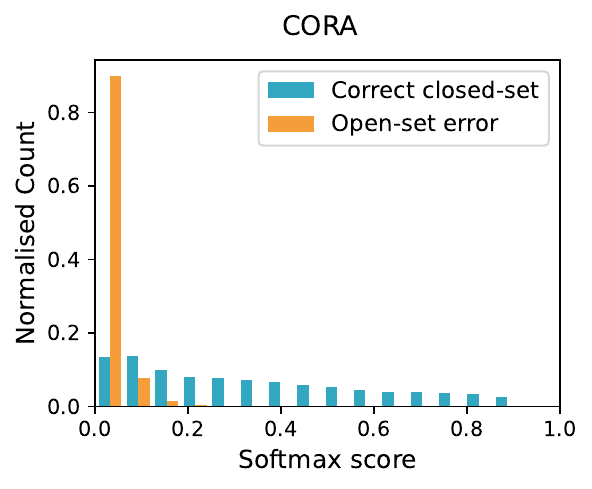}
\end{subfigure}

\caption{The VLM object detectors are prone to underconfident correct closed-set predictions, i.e. correct predictions with low softmax score. The histograms show normalised counts, as there is large imbalance in the magnitude of the correct closed-set and open-set error sets.}
\label{det_unchist}
\end{figure}

In Fig.~\ref{cls_unchist} and Fig.~\ref{det_unchist}, we show histograms visualising prediction uncertainty for the VLM classifiers and detectors respectively. We compare the uncertainty from all correct closed-set predictions with the uncertainty from all open-set errors. In particular, the histograms highlight the different uncertainty behaviours of VLM classifiers and detectors -- while the VLM classifiers are more prone to overconfident open-set errors (i.e. open-set errors with very high softmax score), the VLM detectors are more prone to underconfident correct closed-set predictions (i.e. correct predictions with low softmax score).

\section{Negative Embeddings and Uncertainty Interactions}
\begin{figure}[h!]
    \centering
\includegraphics[width=0.95\linewidth]{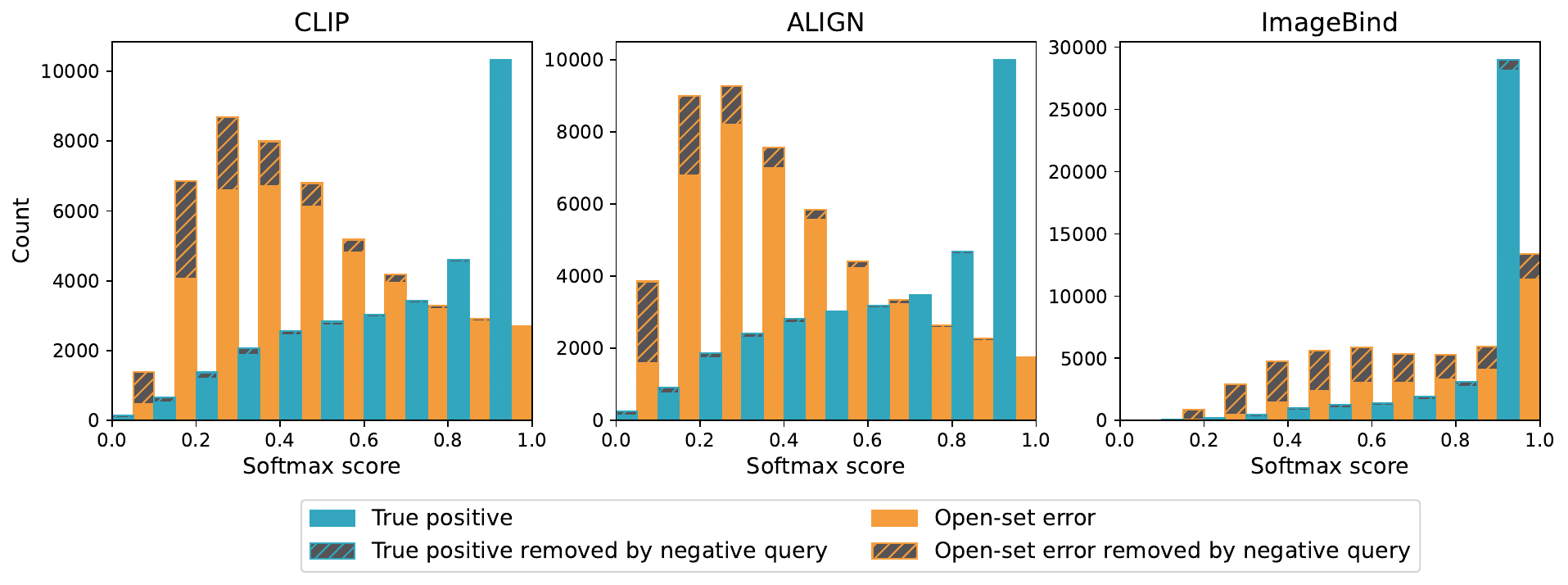}
    \caption{Using negative queries (random 500 words) removes open-set error with different uncertainties depending on the VLM classifier.}
    \label{fig:cls_negquery_removed}
\end{figure}

For some VLM classifiers, the negative queries only capture and remove ``easy'' open-set errors, i.e. open-set errors that already have high uncertainty. See Fig.~\ref{fig:cls_negquery_removed}, which shows the predictions that are removed when introducing $500$ random words as negative queries. In contrast to CLIP and ALIGN, ImageBind is able to remove previously overconfident open-set errors with these negative queries, resulting in up to $5.5\%$ increase in AuPR.

\section{Investigating Individual Random Embedding Efficacy}
Some negative embeddings are more effective for capturing open-set errors. Below, we plot the closed-set versus open-set trade-off for $2500$ random embeddings when used with the VLM classifiers. Ideally, an embedding should capture many open-set images without capturing any closed-set images. Green shading shows embeddings that capture more open-set than closed-set images, and red shading shows the opposite.

\begin{figure}[b!]
\centering
\begin{subfigure}[b]{0.6\textwidth}
  \centering
  \includegraphics[width=\linewidth]{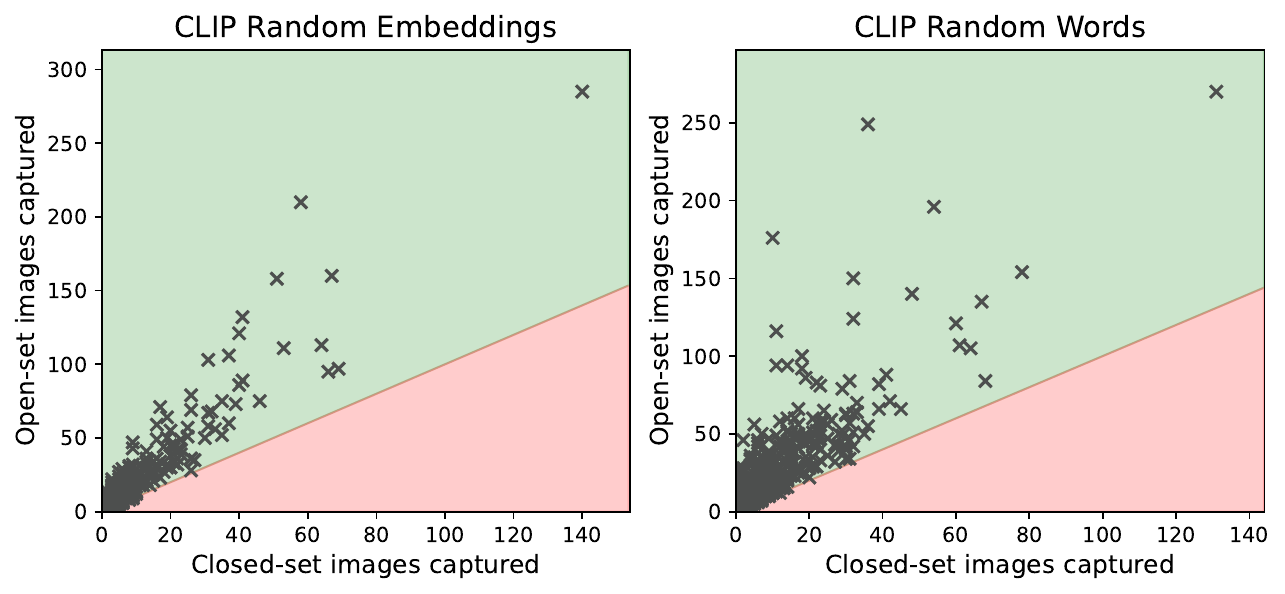}
\end{subfigure}%
\end{figure}
\begin{figure}
\centering
\begin{subfigure}[ht]{0.6\textwidth}
  \centering
  \includegraphics[width=\linewidth]{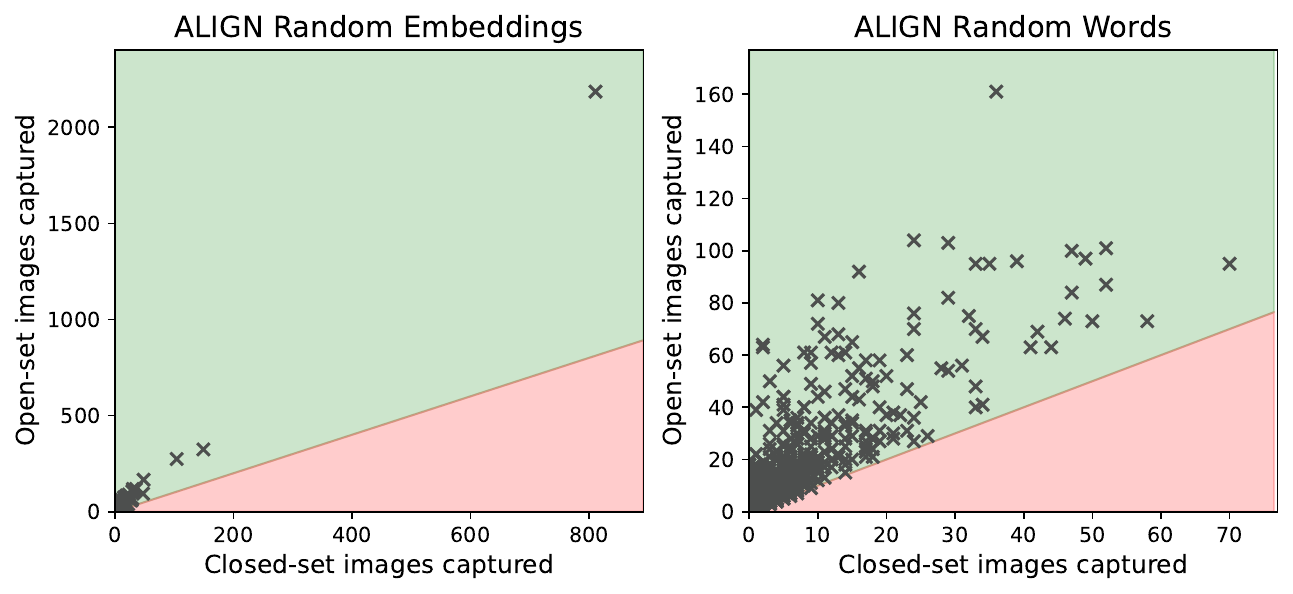}
\end{subfigure}

\begin{subfigure}[b]{0.6\textwidth}
  \centering
\includegraphics[width=\linewidth]{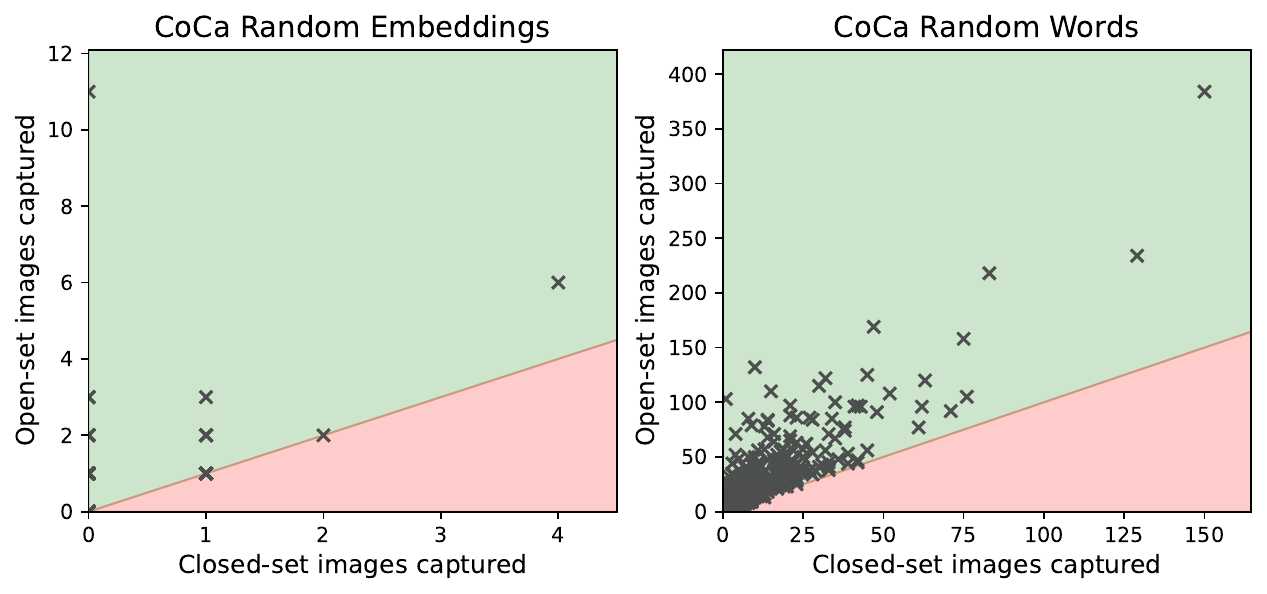}
\end{subfigure}

\begin{subfigure}[b]{0.6\textwidth}
  \centering
  \includegraphics[width=\linewidth]{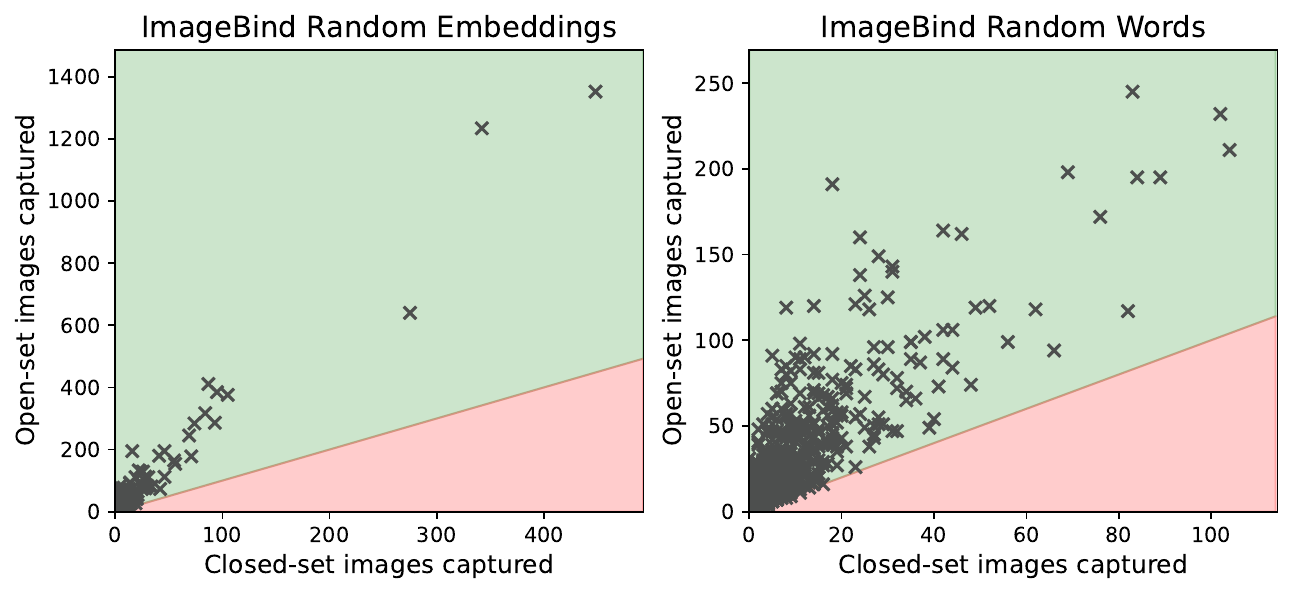}
\end{subfigure}

\begin{subfigure}[b]{0.6\textwidth}
  \centering
  \includegraphics[width=\linewidth]{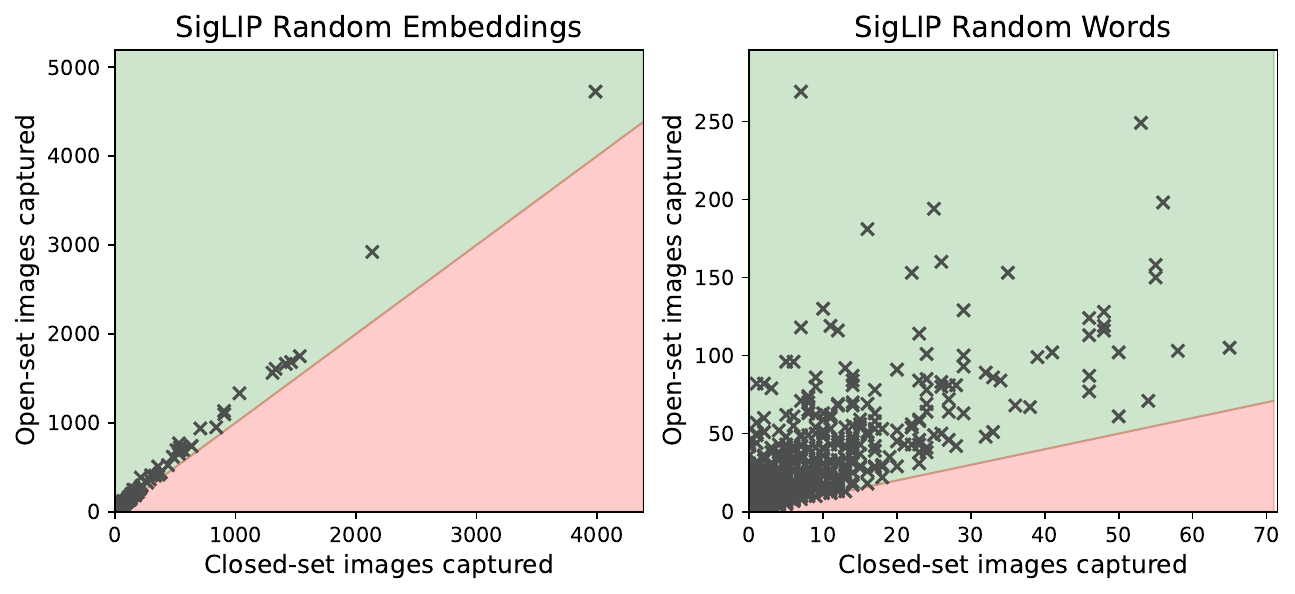}
\end{subfigure}

\begin{subfigure}[b]{0.6\textwidth}
  \centering
  \includegraphics[width=\linewidth]{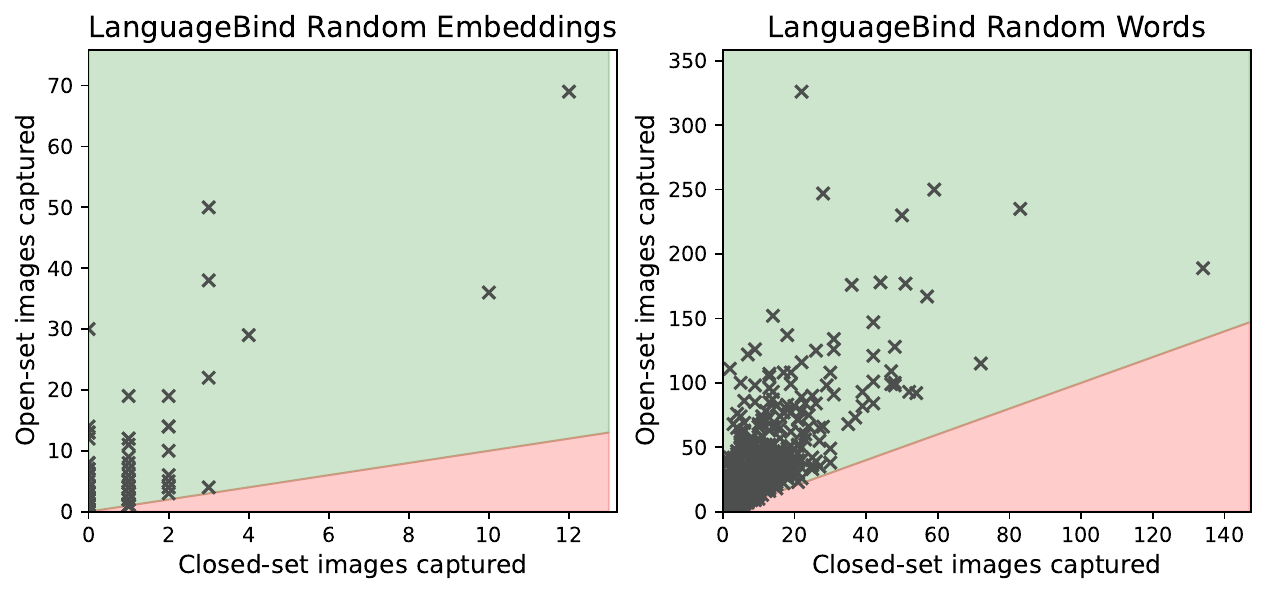}
\end{subfigure}

\caption{Trade-off of individual negative embeddings for capturing closed-set images incorrectly versus capturing open-set images correctly.}
\label{cls_emb_effectiveness}
\end{figure}

\clearpage

%% file: tables/vlm_cls_impl.tex
\begin{table}[th!]
% \caption{}
\setlength{\tabcolsep}{12pt}
\label{tab:vlm_cls_impl}
\centering
% \resizebox{\columnwidth}{!}{%
\begin{tabular}{@{}lcc@{}}
\toprule
                           Classifier   & Repository & Extra Details  \\
\midrule
CLIP~\cite{radford2021clip} & \href{https://github.com/openai/CLIP}{OpenAI CLIP}& ViT-B/32 model \\
\\
\multirow{2}{*}{ALIGN~\cite{jia2021align}} & \multirow{2}{*}{\href{https://github.com/huggingface/transformers/tree/main}{HuggingFace Transformers}} & Kakao Brain implementation\\
& & trained on COYO dataset~\citeSM{kakaobrain2022coyo-700m}. \\
\\

\multirow{2}{*}{CoCa~\cite{yu2022coca}} & \multirow{2}{*}{\href{https://github.com/mlfoundations/open_clip}{OpenCLIP}\citeSM{openclipsoftware}} & ViT-B/32 model\\
& & trained on LAION-2B~\citeSM{schuhmann2022laionb}. \\
\\

ImageBind~\cite{girdhar2023imagebind} & \href{https://github.com/facebookresearch/ImageBind}{Meta AI ImageBind} & imagebind\_huge\\
\\

\multirow{1}{*}{SigLIP~\cite{siglip}} & \multirow{1}{*}{\href{https://github.com/huggingface/transformers/tree/main}{HuggingFace Transformers}} &  google/siglip-base-patch16-224\\
\\
\multirow{1}{*}{LanguageBind~\cite{languagebind}} & \multirow{1}{*}{\href{https://github.com/PKU-YuanGroup/LanguageBind}{LanguageBind Repository}} & LanguageBind\_Image\\

                            \bottomrule
\end{tabular}
% }
\end{table}

%% file: tables/vlm_det_impl.tex
\begin{table}[th!]
% \caption{}
\setlength{\tabcolsep}{12pt}
\label{tab:vlm_cls_impl}
\centering
% \resizebox{\columnwidth}{!}{%
\begin{tabular}{@{}lcc@{}}
\toprule
                           Detector   & Repository & Config file or testing script\\
\midrule
OVR-CNN~\cite{ovrcnn} & \href{https://github.com/alirezazareian/ovr-cnn}{Official Paper Repo} & zeroshot\_v06.yaml \\
\\
ViLD~\cite{vild} & \href{https://github.com/tensorflow/tpu/tree/master/models/official/detection/projects/vild}{Official Paper Repo}
& ViLD\_demo.ipynb \\
& & \texttt{FLAGS.use\_softmax} set to \texttt{True} \\
\\
OV-DETR~\cite{ovdetr} & \href{https://github.com/yuhangzang/OV-DETR}{Official Paper Repo} & Default evaluation setup \\
& & shown in \texttt{run\_scripts.md}\\
\\
RegionCLIP~\cite{regionclip} & \href{https://github.com/microsoft/RegionCLIP}{Official Paper Repo} & RN50, COCO Configuration \\
& & (Generalized: Novel + Base) \\
& & in \texttt{test\_transfer\_learning.sh} \\
\\
Detic~\cite{detic} & \href{https://github.com/facebookresearch/Detic}{Official Paper Repo} & Detic\_OVCOCO\_CLIP\_\\
& & R50\_1x\_max-size\_caption.yaml\\
\\
VLDet~\cite{vldet} & \href{https://github.com/clin1223/VLDet}{Official Paper Repo} & VLDet\_OVCOCO\_CLIP\_\\
& & R50\_1x\_caption\_custom.yaml \\
\\
CORA~\cite{cora} & \href{https://github.com/tgxs002/CORA}{Official Paper Repo} & R50\_dab\_ovd\_3enc\_\\
& & apm128\_splcls0.2\_relabel\_noinit.sh\\
\\

                            \bottomrule
\end{tabular}
% }
\end{table}

%% file: tables/detection_coco_auroc.tex
\begin{table}[th!]
\caption{Comparison between AuPR and AuROC metrics for the state-of-the-art open-vocabulary object detectors (with ResNet backbone indicated), tested on COCO. Arrows indicate the direction of optimal performance.}
\label{tab:det_auroc}
\centering
\resizebox{\columnwidth}{!}{%
\begin{tabular}{@{}lccccccc@{}}
\toprule
                            &     & Negative & AuPR      &  AuROC & TP & OSE  \\
Detector  & Uncertainty & Embedding & \multicolumn{1}{c}{$\uparrow$} & $\uparrow$ & $\uparrow$    & $\downarrow$         &        \\ \midrule
\multirow{3}{*}{OVR-CNN (R50) \cite{ovrcnn}}    & Softmax  & Zero-Emb  & 75.4    & 93.0 & \multirow{3}{*}{13,544} & \multirow{3}{*}{190,146}   \\
    & Cosine  & Zero-Emb & \multicolumn{1}{c}{77.2}       & 94.1 & \\
    & Entropy  & Zero-Emb   & \multicolumn{1}{c}{63.2}       & 81.2 & \\ 
\midrule
\multirow{3}{*}{ViLD (R152)\cite{vild}}        & Softmax  & ``background''   & 60.1 & 97.0 & \multirow{3}{*}{16,011} & \multirow{3}{*}{1,485,600} & \\
 & Cosine & ``background'' & 33.1 & 96.7  &                          \\
 & Entropy & ``background'' & 62.2 & 97.3 &                         \\ \midrule
\multirow{1}{*}{OV-DETR (R50) \cite{ovdetr}}     & Sigmoid  & None  &    75.8      & 97.4 & 15,818 & 1,485,600 \\ \midrule

\multirow{3}{*}{RegionCLIP (R50)\cite{regionclip}}  & Softmax & Zero-Emb    &   73.1  &  95.4 &    \multirow{3}{*}{15,741}    & \multirow{3}{*}{493,940}  \\
& Cosine  & Zero-Emb    &  74.0  & 94.3  &  \\
& Entropy & Zero-Emb    & 35.4 & 75.9 &                         \\ \midrule
\multirow{1}{*}{Detic (R50)\cite{detic}}      & Sigmoid & Zero-Emb  &      72.6    &  95.0  &     14,670                 &    495,200    \\ \midrule
\multirow{1}{*}{VLDet (R50)\cite{vldet}}       & Sigmoid     & Zero-Emb     &   79.8      & 92.8 &      14,751                &   112,923                    \\ \midrule
\multirow{1}{*}{CORA (R50)\cite{cora}}       & Sigmoid     &  Zero-Emb   & 67.2    &  95.0 &  13,763                    &   495,200                   \\ 

\bottomrule

\end{tabular}
}
\end{table}

%% file: tables/classification_csvsos.tex
\begin{table}[h]
\caption{Difference in uncertainty (Softmax or Sigmoid) performance between open-set error and closed-set error identification for the VLM classifiers tested on ImageNet1k.}
\label{tab:cls_unc_csvsos}
\centering
\resizebox{0.8\columnwidth}{!}{%
\begin{tabular}{@{}lcccccccc@{}}
\toprule
                            &    & AuROC & AuPR      &  P@95R & R@95P & TP & Error Count  \\
Classifier  & Error & \multicolumn{1}{c}{$\uparrow$} & $\uparrow$ & $\uparrow$ & $\uparrow$ & $\uparrow$     & $\downarrow$              \\  \midrule
\multirow{1}{*}{CLIP \cite{radford2021clip}}  & Open-set & 79.7 &   71.5  &  46.2  & 3.4  &    \multirow{1}{*}{31,023}    & \multirow{1}{*}{50,000} \\
(Softmax) & Closed-set   &  80.7 & 87.4 & 68.5 & 30.0 & 31,023 & 18,977 \\

                            \midrule
\multirow{1}{*}{ALIGN \cite{jia2021align}}  & Open-set & 81.0 & 74.7  &  48.1  & 8.3  &    \multirow{1}{*}{32,618}    & \multirow{1}{*}{50,000} \\
(Softmax) & Closed-set   & 81.0 & 89.2 & 70.9 & 35.7 & 32,618 & 17,382 \\
                       \midrule
\multirow{1}{*}{CoCa \cite{yu2022coca}} & Open-set & 79.0 & 71.7 & 46.0 & 3.4 & \multirow{1}{*}{31,725} & \multirow{1}{*}{50,000}\\
(Softmax) & Closed-set & 81.8 & 88.8 & 70.3 & 33.1 & 31,725 & 18,275\\

                            \midrule
\multirow{1}{*}{ImageBind \cite{girdhar2023imagebind}}  & Open-set & 82.8 & 79.1  & 52.8  & 5.0  &    \multirow{1}{*}{38,405}    & \multirow{1}{*}{50,000} \\
(Softmax) & Closed-set  &  83.4 & 94.1 & 82.3 & 57.1 & 38,405 & 11,595  \\
 
\midrule
\multirow{1}{*}{SigLIP \cite{siglip}} & Open-set & 81.0 & 76.2 & 52.1 & 2.2 & 37,851 & 50,000  \\
(Sigmoid) & Closed-set & 70.2 & 87.1 & 78.2 & 5.1 & 37,851 & 12,149 \\

                            \midrule
\multirow{1}{*}{LanguageBind \cite{languagebind}} & Open-set & 83.9 & 80.7 & 54.7 & 10.5 & \multirow{1}{*}{39,243} & \multirow{1}{*}{50,000}\\
(Softmax) & Closed-set & 84.2 & 94.8 & 84.1 & 63.3 & 39,243 & 10,757\\
                      
                            \bottomrule
\end{tabular}
}
\end{table}

%% file: tables/detection_csvsos.tex
% \usepackage{multirow}

\begin{table}[h!]
\caption{Difference in uncertainty performance between open-set error and closed-set error identification for the VLM detectors tested on COCO.}
\label{tab:det_unc_csvsos}
\centering
\resizebox{0.8\columnwidth}{!}{%
\begin{tabular}{@{}lccccccc@{}}
\toprule
                            &      & AuPR      &  P@95R & R@95P & TP & Error Count                                    \\
Detector  & Error  & \multicolumn{1}{c}{$\uparrow$} & $\uparrow$ & $\uparrow$ & $\uparrow$     & $\downarrow$                  \\ \midrule
\multirow{1}{*}{OVR-CNN (R50) \cite{ovrcnn}}    & Open-set  & 75.4    & 13.0 & 42.9  & \multirow{1}{*}{13,544} & \multirow{1}{*}{190,146} \\
  (Softmax)  & Closed-set & 81.8 & 25.7 & 50.9 & 13,544 & 68,708 \\
  
\midrule
\multirow{1}{*}{ViLD (R152)\cite{vild}}        & Open-set   & 60.1 & 7.5 & 15.4 & \multirow{1}{*}{16,011} & \multirow{1}{*}{1,485,600}  \\
 (Softmax) & Closed-set & 75.6 & 28.0 & 29.7 & 16,011 & 126,774 \\
 \midrule
 
\multirow{1}{*}{OV-DETR (R50) \cite{ovdetr}} & Open-set &    75.8      & 5.9  & 47.3 & 15,818 & 1,485,600 \\ 
(Sigmoid) & Closed-set & 78.1 & 9.0 & 48.2 & 15,818 & 461,950 \\ \midrule

\multirow{1}{*}{RegionCLIP (R50)\cite{regionclip}}  & Open-set    &   73.1  &  9.9  & 39.4  &    \multirow{1}{*}{15,741}    & \multirow{1}{*}{493,940} \\
(Softmax) & Closed-set & 81.6 & 24.1 & 53.4 & 15,741 & 120,483 \\
\midrule

\multirow{1}{*}{Detic (R50)\cite{detic}}   &   Open-set  &      72.6    &   8.1      & 42.2     &     14,670                 &    495,200                   \\
(Sigmoid) & Closed-set & 80.4 & 20.1 & 52.4 & 14,670 & 154,824\\ \midrule

\multirow{1}{*}{VLDet (R50)\cite{vldet}}       & Open-set  &   79.8      &  21.9    & 48.5 &      14,751                &   112,923\\
(Sigmoid) & Closed-set & 82.8 & 29.2 & 52.0 & 14,751 & 64,247 \\ \midrule

\multirow{1}{*}{CORA (R50)\cite{cora}}       & Open-set & 67.2    &    8.1    & 24.7   &  13,763                    &   495,200 \\
(Sigmoid) & Closed-set & 77.4 & 20.0 & 45.8 & 13,763 & 101,015 \\ 

\bottomrule

\end{tabular}
}
\end{table}